\documentclass{IEEEtran}
\usepackage{cite}
\usepackage{amsmath,amssymb,amsfonts}
\usepackage{algorithmic}
\usepackage{algorithm}
\usepackage{graphicx,dblfloatfix}
\usepackage{textcomp}
\def\BibTeX{{\rm B\kern-.05em{\sc i\kern-.025em b}\kern-.08em
    T\kern-.1667em\lower.7ex\hbox{E}\kern-.125emX}}
    
\usepackage{booktabs}
\usepackage{subfigure}
\usepackage{tabularx}
\usepackage{color}
\usepackage{ragged2e}
\newcolumntype{Y}{>{\RaggedRight\arraybackslash}X} 
\usepackage{balance}

\newcommand{\degree}{$^{\circ}$}

\title{\LARGE \bf
   OpenSHC: A Versatile Multilegged Robot Controller
}

\author{Benjamin Tam$^{\dagger\ddagger}$, Fletcher Talbot$^{\dagger}$, Ryan Steindl$^{\dagger}$, Alberto Elfes$^{\dagger}$, Navinda Kottege$^{\dagger}$
\thanks{$^{\dagger}$Robotics and Autonomous Systems
Group, Commonwealth Scientific and Industrial Research Organisation (CSIRO), Pullenvale, QLD 4069, Australia. This work was fully funded by the CSIRO. Correspondence should be addressed to  
        {\tt\small navinda.kottege@csiro.au}}\\
\thanks{$^{\ddagger}$School of Information Technology and Electrical Engineering, The University of Queensland, St. Lucia, QLD 4072, Australia.}        
}

\begin{document}
	
\maketitle
\thispagestyle{empty}
\pagestyle{empty}

\begin{abstract}

Multilegged robots have the ability to perform stable locomotion on relatively rough terrain. However, the complexity of legged robots over wheeled or tracked robots make them difficult to control. This paper presents OpenSHC (Open-source Syropod High-level Controller), a versatile high-level controller {capable of generating gaits and poses for quasi-static multilegged robots, both simulated and with real hardware implementations. With full Robot Operating System (ROS) integration, the controller can be quickly deployed on robots with different actuators and sensor payloads}. The flexibility of OpenSHC is demonstrated on the 30 degrees of freedom hexapod Bullet, analysing the energetic performance of various leg configurations, kinematic arrangements and gaits over different locomotion speeds. With OpenSHC being easily configured to different physical and locomotion specifications, a hardware-based parameter space search for optimal locomotion parameters is conducted. The experimental evaluation shows that the mammalian configuration offers lower power consumption across a range of step frequencies; with the insectoid configuration providing performance advantages at higher body velocities and increased stability at low step frequencies. OpenSHC is open-source and able to be configured for various number of joints and legs.

\end{abstract}

\maketitle

\section{Introduction}
\label{sec:introduction}

\begin{figure*}[!t]
    \centering
    \includegraphics[width=170mm]{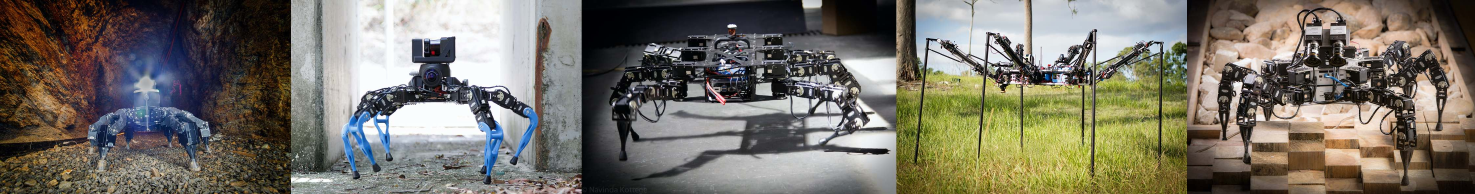}
    \caption{A number of different multilegged robots that use OpenSHC, from left to right: Gizmo, Zee, Bullet, MAX and Weaver. 
    }
\label{fig:hexapodbanner}
\end{figure*}

Legged robots have advantages compared to their wheeled and tracked counterparts when navigating in complex terrain. From their ability to traverse discontinuous terrain, climb over obstacles and disturb the terrain minimally, to probing the terrain and manipulating the environment without an additional arm \cite{Roennau2014,tennakoon_safe_2020b}. However, the trade-off of this versatility and mobility is the significant challenges in mechanical and control complexity \cite{Hutter2016}. The combination of both the robot platform design and the control algorithm determines the performance and effectiveness of the robot. To further research and development of legged robots, the authors present OpenSHC - Open-source Syropod High-level Controller - a versatile controller that is capable of generating statically stable gaits for multilegged robots\footnote{https://github.com/csiro-robotics/syropod\_highlevel\_controller}. It is the result of legged robot locomotion research conducted at CSIRO's Robotics and Autonomous Systems Group since 2011, with some of the robots running OpenSHC shown in Fig.~\ref{fig:hexapodbanner}. {Using the Robot Operating System (ROS) framework for modularity and easy deployment, OpenSHC is designed to generate foot tip trajectories for a given gait sequence, step clearance, step frequency and input body velocity for many different legged robots with various leg configurations and degrees of freedom. Any legged robot with up to 8 legs that can be specified using Denavit-Hartenberg (DH) parameters} \cite{Hartenberg1964} {can be used with OpenSHC. Input sensors such as IMUs and joint effort feedback can be utilised by the controller to provide robust trajectories in inclined and uneven terrain. OpenSHC can also be used in simulation to control robots in rviz and Gazebo simulation environments. This provides a convenient way for roboticists to design legged robots by being able to tune various parameters to fit specific performance criteria.}

In order to show the capability of OpenSHC, a study on locomotion efficiency of a 30 degrees of freedom (DOF) hexapod robot platform called \textit{Bullet} in the nature inspired mammalian and insectoid (sprawling-type) configuration is presented. Six legged robots have an advantage over bipeds and quadrupeds when it comes to statically stable locomotion on mild terrain, where the fast tripod gait has an energy efficiency advantage over other gaits \cite{kottege2015energetics}. In the mammalian configuration, the legs are below the body, reducing the support polygon while decreasing the power consumption required to support the body. Insectoid configuration places the legs to the side of the body, lowering the centre of mass, increasing locomotion workspace and stability. The different physical and locomotion specifications require unique parameters for control, something OpenSHC allows for easily. The experimental results provide unique insights into the novel parameter space of hardware changes to leg arrangement and configuration; and locomotion changes to step frequency, stride length and gait.

The design philosophy and history behind OpenSHC is presented in Section~\ref{sec:designphilosophy}. The kinematic algorithms in the controller is summarised in Section~\ref{sec:kinematicmodel}. Section~\ref{sec:systemarchitecture} provides an overview of OpenSHC with details of each sub-component of the system. Section~\ref{sec:casestudy} describes the study of locomotion efficiency with respect to popular leg configurations in literature, while Section~\ref{sec:robotplatform} describes the mechanical specifications of Bullet. Experiments are explained in Section~\ref{sec:experiments} with results shown in Section~\ref{sec:results} and then discussed in Section~\ref{sec:discussion}. Section \ref{sec:conclusions} concludes the paper and provides areas of focus for the future of OpenSHC.


\begin{figure*}[b!]
\centering
    \subfigure[]{\label{fig:gazebo_bullet} \includegraphics[height=3.3cm]{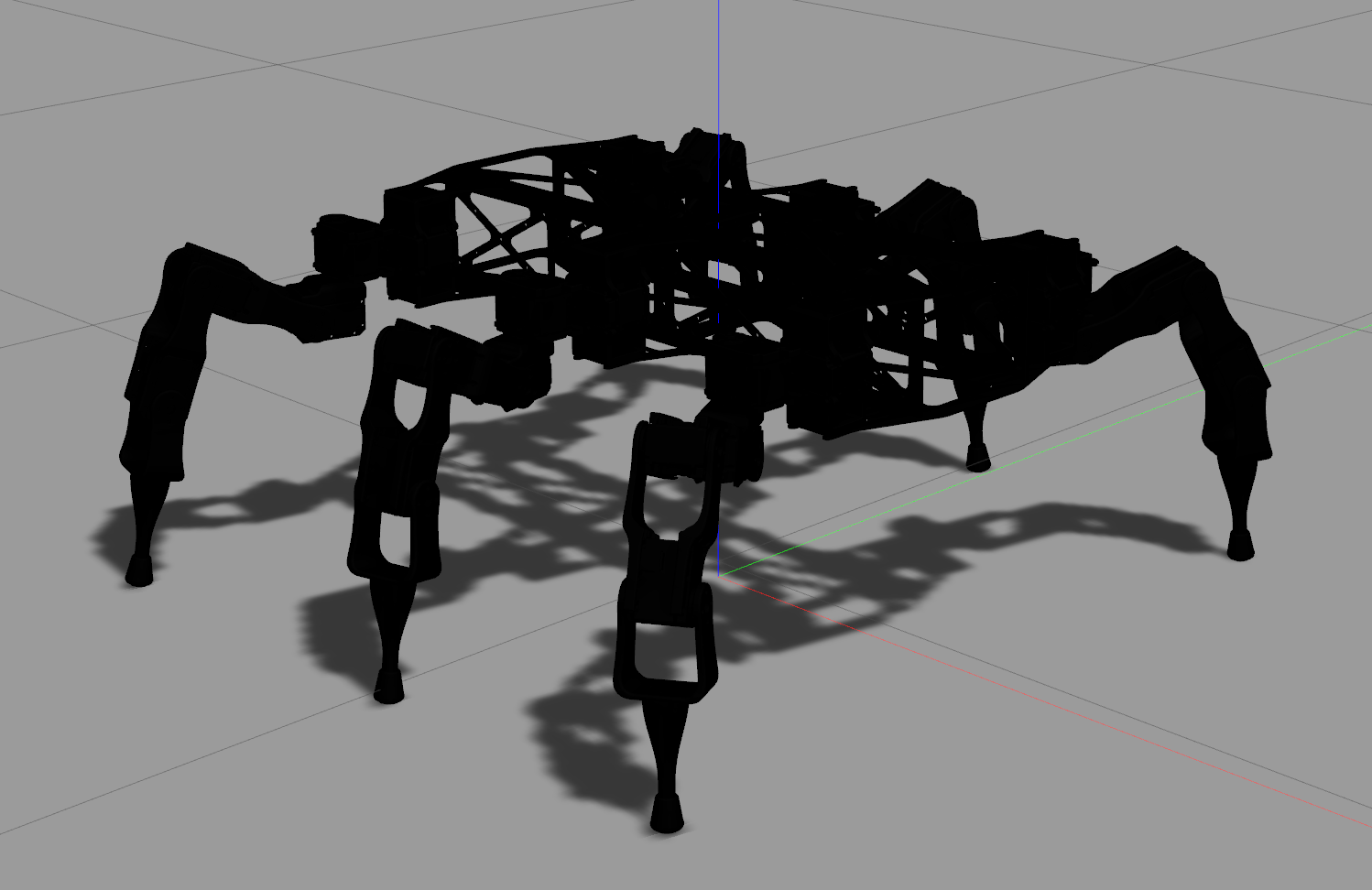}}
    \subfigure[]{\label{fig:gazebo_max} \includegraphics[height=3.3cm]{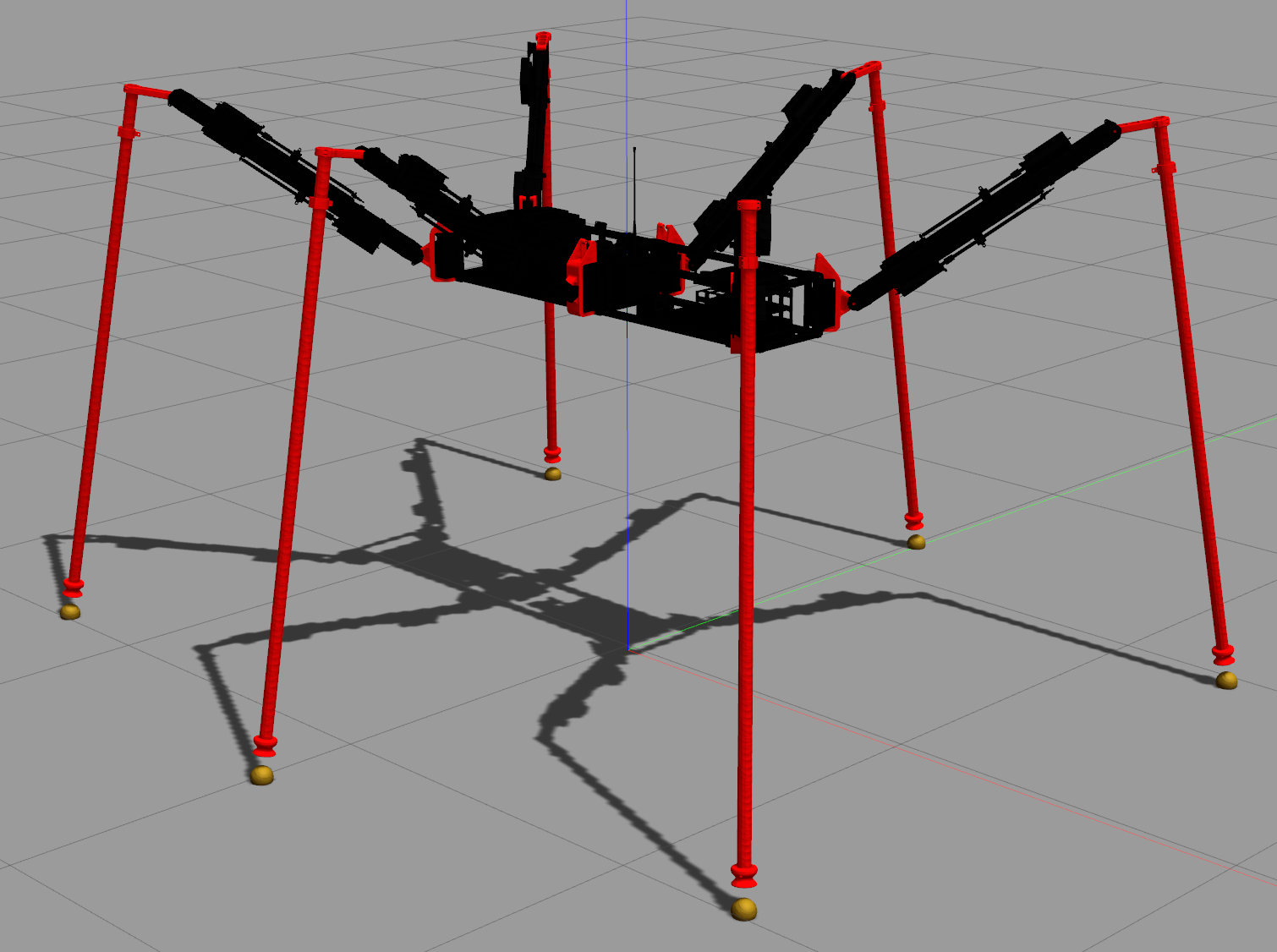}}
    \subfigure[]{\label{fig:gazebo_weaver2} \includegraphics[height=3.3cm]{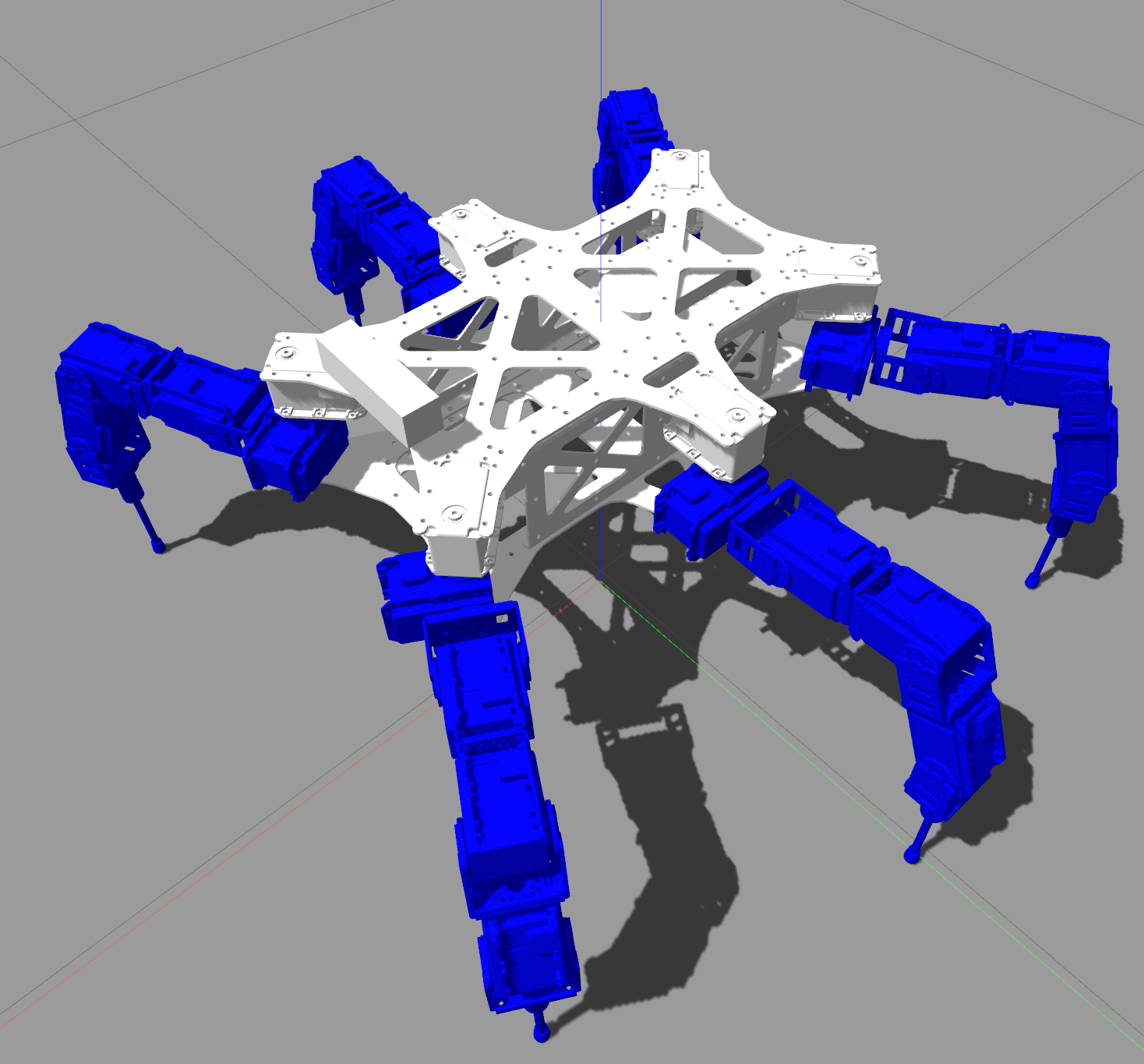}}
    \subfigure[]{\label{fig:gazebo_magneto} \includegraphics[height=3.3cm]{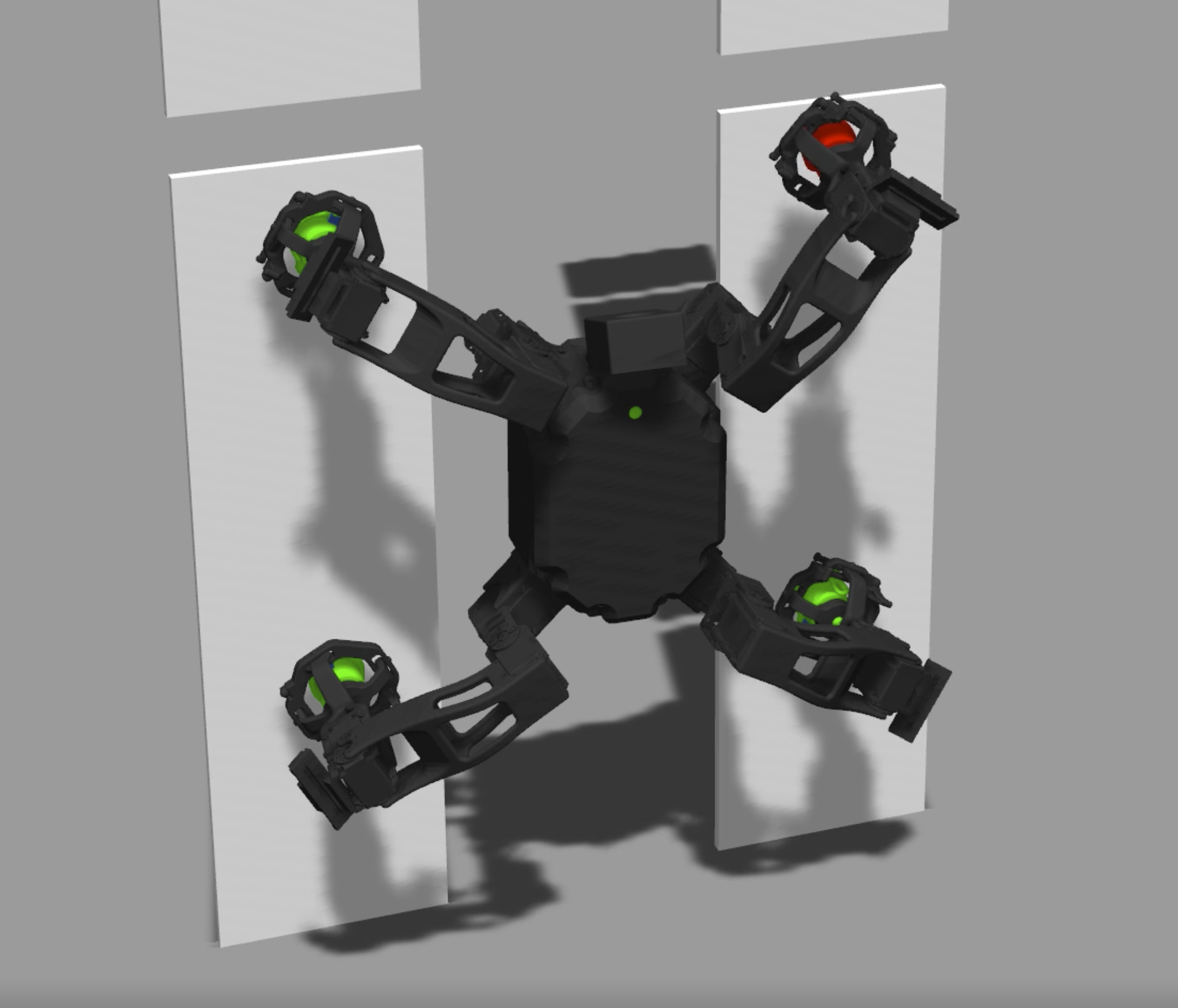}}

\caption{{OpenSHC used for simulating multilegged robots in Gazebo, from left to right: Bullet, MAX, Weaver and Magneto.}}
\label{fig:gazeborobots}
\end{figure*}

\section{Design Philosophy}
\label{sec:designphilosophy}
{The design philosophy of OpenSHC was to create a modular controller for research and development on different simulated and real hardware robots without having to redevelop the controller to make each robot walk. }

{A variety of open-source multilegged robot projects exist, each focusing on different challenges and applications. Projects such as OpenRoACH} \cite{8794042}, {Open Dynamic Robot Initiative} \cite{grimminger2020open} {and Oncilla} \cite{Sprowitz2018} {have focused on developing a complete robot platform with mechanical and electrical hardware designs, and control software that is tightly coupled to the hardware specifications. OpenSHC on the other hand, is developed to be applicable to many different legged robots including those in simulation. We also differ from the Phoenix} \cite{phoenix} {robot control software used to control Trossen Robotics PhantomX hexapods} \cite{trossen}{, through full ROS} \cite{quigley_2009} {integration, the de facto robotics middleware used by the research community and industry.}

{OpenSHC is a controller that is not linked to a particular hardware design. That is, OpenSHC allows robot morphology to be iteratively tested in simulation first, before being deployed onto a real system.}

{Through its full ROS integration, OpenSHC has the ability to easily interface with simulated robots in Gazebo, the ROS compatible 3D simulation engine, using its Gazebo control interface. This allows testing of various features and design parameters in designing a robot without having to go through hardware based iterations. This is done  through the use of a Universal Robot Description Format (URDF) file which describes the robot model's links, joints, simulated actuators and IMU. The URDF file for the robot platform Bullet, used in the case study in} Section~\ref{sec:casestudy} {is provided as part of the suite of OpenSHC open source software packages. Using OpenSHC to command the joints of a robot model, a user can test custom robot morphologies, gaits, stance positions and various other parameters to optimise a robot design and test it in a full physics simulated environment. In development of the hexapod robot Bruce, used in the DARPA SubT Challenge} \cite{steindl_2020}, {initial analysis of the actuator requirements was undertaken using OpenSHC and an initial design model of Bruce with simulated actuators in Gazebo. This analysis involved commanding the initial Bruce model to run at a variety of velocities and gaits, with a variety of payload weights and recording the simulated actuator torque requirements. Those requirements were used to inform the final design of Bruce including morphology and choice of actuator. Legged robots such as MAX} \cite{Elfes2017}, {a 2.25\,m 18\,DOF hexapod; Weaver} \cite{Bjelonic2016} {and Bullet} \cite{tennakoon_safe_2020b}, {both 30\,DOF hexapod robots; and Magneto} \cite{bandyopadhyay_2018}, {a quadruped with 3\,DOF actuated limbs and 3\,DOF compliant magnetic feet have also been extensively simulated using OpenSHC to develop new applications and functionality} (Fig.~\ref{fig:gazeborobots}). 

The following list highlights some of the key features of OpenSHC:
\begin{itemize}
\item Fully configurable for a variety of platform designs with differing physical characteristics, including up to 8 legs each with up to 6\,DOF per leg.
\item Dynamically switchable gait options with ability to design custom gaits.
\item User defined body clearance, step clearance and step frequency.
\item Manual body posing in 6\,DOF.
\item Manual leg manipulation (legipulation) for up to two legs simultaneously with toggle of manipulation control of either tip position in Cartesian space or direct control of joint positions (3\,DOF legs only).
\item Startup direct mode to move foot tip positions linearly from initial position to default walking stance positions.
\item Startup sequence mode with full chain of startup/shutdown sequences to start from a `packed' state and generate a sequence to stand up off the ground into its default walking stance; and similarly able to shutdown and transition back to a packed state.
\item Cruise control mode to set robot velocity as a constant predefined input velocity or set to the current input velocity.
\item Auto navigation mode when interfaced with high level navigation stack.
\item Optional admittance control with dynamic leg stiffness to ensure leg contact with ground and offer moderately rough terrain walking ability.
\item Optional IMU body compensation to keep body horizontally level at all times, using IMU data.
\item Optional inclination compensation which strives to keep body centre of gravity over the estimated centroid of the support polygon whilst walking on inclined planes.
\item Optional bespoke automatic body posing system to pose each robot leg cyclically as defined by auto-pose parameters.
\end{itemize}

A variety of research projects conducted in our lab has contributed and benefited from the controller. {Novel functionality developed on particular robot platforms are generalised and incrementally added into the controller so other platforms can benefit from the work.  Different robot platforms such as MAX, Weaver and Magneto mentioned earlier, have influenced the design and functionality of the controller. The simple command interface for sending velocity control to the system enabled higher level autonomy to be integrated and tested onto hardware easily. Research in probing for brittle terrain} \cite{tennakoon_safe_2020b}, steep terrain ascent \cite{molnar_2017}, adapting robot pose for confined spaces \cite{buchanan_walking_2019}, augmented telepresence for remote inspection \cite{tam_2017} and autonomous adaptation of locomotion parameters \cite{Homberger2016,Bjelonic2017,bjelonic_weaver:_2018} have all built upon the OpenSHC framework.

\section{Kinematic Model}
\label{sec:kinematicmodel}
{In OpenSHC, the robot is represented as a kinematic model within the controller using the Denavit-Hartenberg (DH) parameters} \cite{Hartenberg1964,Spong2006}. {This allows for the joint angles and end effector locations to be easily transformed for forward kinematics (FK) and inverse kinematics (IK) calculations. The robot's body frame is represented as pose $(o_bx_by_bz_b)$, consisting of the body orientation $(o_{roll}, o_{pitch}, o_{yaw})$ and displacement in $(x,y,z)$. This robot frame provides the odometry of the robot to a reference map frame $(o_mx_my_mz_m)$ which is fixed to the world. The coordinate frames are right-handed with $x$ forward, $y$ left and $z$ up. Each leg has its own leg frame $(o_1x_1y_1z_1)$, which is the static offset from the robot's body frame origin to the centre of rotation of the first joint of that leg, when moving outward from the robot centre. The leg frame is used to transform the desired robot body movement to the corresponding leg motions via FK and IK. }

{The leg number convention used in OpenSHC (shown in} Fig.~\ref{fig:bulletstructure}) {follows a clockwise sequence from the front right leg as `1'} \cite{belter_2010}. {This convention facilitates arbitrary body designs such as elongated or axis-symmetric circular bodies} \cite{clemente2011gait}. {Joint and link names are bio-inspired from insect morphology} \cite{Nelson1997}. {The links are named coxa, femur, tibia and tarsus and the joints immediately before the link is given the same name as the link for easy reference (E.g. tibia joint is between the femur and tibia links). Where there are compound joints with multiple degrees of freedom, the individual degrees of freedom is used as a subscript with the relevant link and joint names (E.g.} coxa\textsubscript{yaw}, coxa\textsubscript{roll}).

\subsection{Forward Kinematics}
\label{sec:forwardkinematics}
The specified DH parameters $\theta_{i}$, $d_{i}$, $a_{i}$ and $\alpha_{i}$ are the rotation around $z$, translation along $z$, translation along $x$ and rotation around $x$, respectively \cite{Waldron2016}. The representation of the combined homogeneous transform of the parameters becomes:
\begin{equation}
    \begin{aligned}
        H_{i+1}^{i} &= Rot_{z,\theta_{i}} \cdot Trans_{z,d_{i}} \cdot Trans_{x,a_{i}} \cdot Rot_{x,\alpha_{i}} \\
            &= \begin{bmatrix} 
            c_{\theta_{i}} & -s_{\theta_{i}}c_{\alpha_{i}} & s_{\theta_{i}}s_{\alpha_{i}} & a_{i}c_{\theta_{i}} \\
            s_{\theta_{i}} & c_{\theta_{i}}c_{\alpha_{i}} & -c_{\theta_{i}}s_{\alpha_{i}} & a_{i}s_{\theta_{i}} \\
            0 & s_{\alpha_{i}} & c_{\alpha_{i}} & d_{i} \\
            0 & 0 & 0 & 1
            \end{bmatrix}
    \end{aligned} \label{eq:DHTransform}
\end{equation}
where $c_{x}$ and $s_{x}$ denote $\cos(x)$ and $\sin(x)$ respectively. {Details on the transformation matrices are provided in Appendix~\ref{sec:FKderivation}. For further information about FK, please see \cite{Waldron2016}.}

\begin{figure}[t!]
    \centering
    \includegraphics[width=75mm]{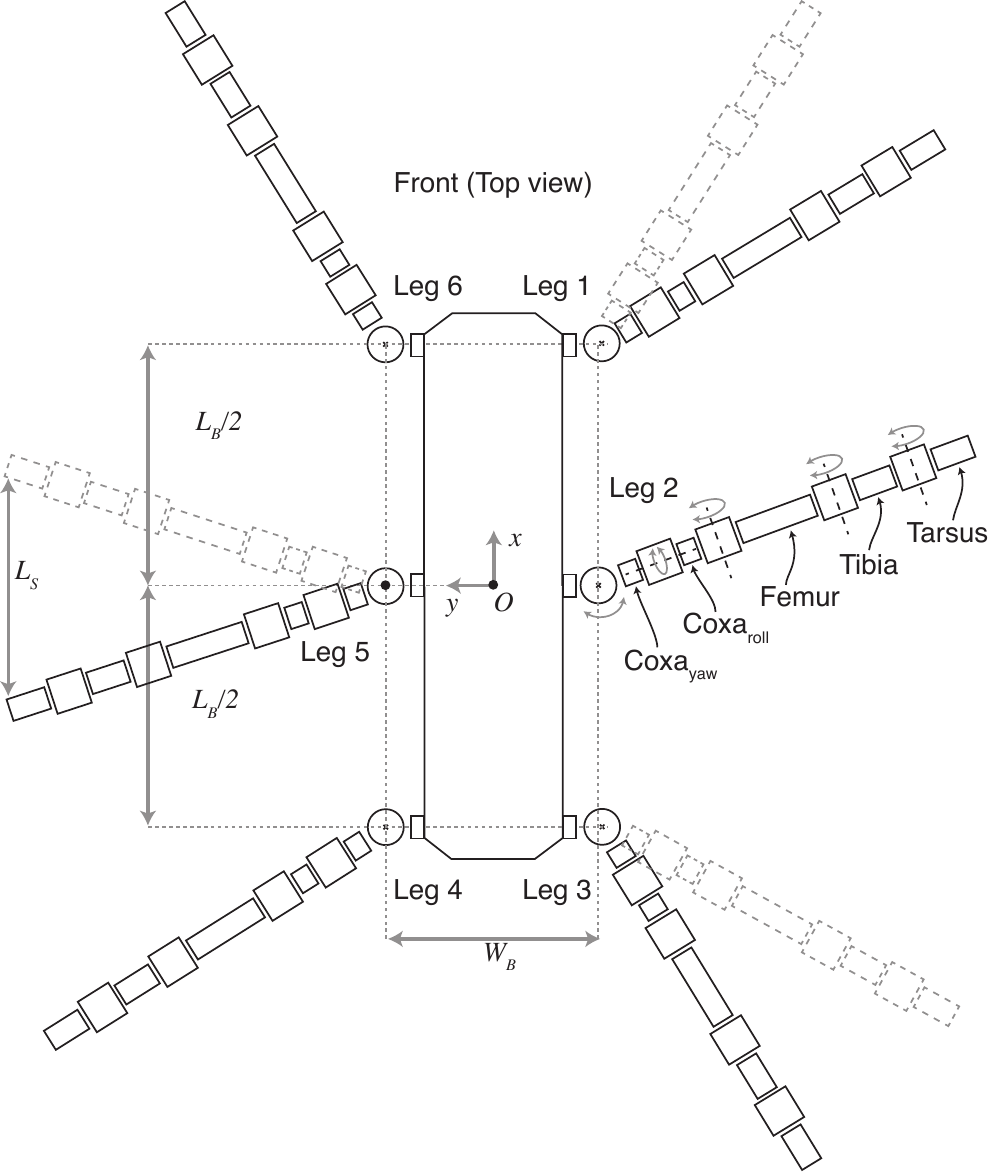}
    \caption{The kinematic structure of the hexapod Bullet. 
    }
\label{fig:bulletstructure}
\end{figure}

 In the notation $H_{i+1}^{i}$, the superscript denotes the reference frame $(o_ix_iy_iz_i)$ and the subscript indicates the transformed frame $(o_{i+1}x_{i+1}y_{i+1}z_{i+1})$. Using Fig.~\ref{fig:bulletstructure} as a 30\,DOF hexapod example, for a leg of the robot the transform from the leg frame $(o_1x_1y_1z_1)$ to the end effector $(o_ex_ey_ez_e)$ is given by:
\begin{equation}
H_{e}^{1}=H_{2}^{1}(q_{1}) \cdot H_{3}^{2}(q_{2}) \cdot H_{4}^{3}(q_{3}) \cdot H_{5}^{4}(q_{4}) \cdot H_{e}^{5}(q_{5})
\label{eq:FKtransform}
\end{equation}
where $q_1,q_2,q_3,q_4$ and $q_5$ are the joint angles for the $\text{coxa}_{\text{yaw}}$, $\text{coxa}_{\text{roll}}$, femur, tibia and tarsus joints respectively.

\subsection{Inverse Kinematics}
\label{sec:inversekinematics}

The IK for calculating the desired change in joint angles ($\Delta \theta$) for a given incremental change in end effector position ($\Delta \vec{\mathbf{s}}$) is calculated using the Jacobian matrix $J$ and the Levenberg-Marquardt method, also known as the damped least squares method. {Full derivation is provided in Appendix~\ref{sec:IKderivation} (based on \cite{buss_2004}), with the resultant equation to solve given by:}

\begin{equation}
    \Delta \theta = J^T\left( J J^T + \lambda^2I\right)^{-1}  \Delta \vec{\mathbf{s}} = \mathbf{Z} \Delta \vec{\mathbf{s}}
    \label{eq:dla}
\end{equation}
where $\lambda > 0\in \mathbb{R}$ and $ \mathbf{Z} = J^T\left( J J^T + \lambda^2I\right)^{-1} $.

For robots with redundant DOFs, joint position/velocity limit avoidance (JLA) strategies are used to increase the safe operation of the joints \cite{fahimi_redundant_2009}. A full-time kinematic optimisation method of JLA is used to keep the joints away from joint position and velocity limits as much as possible. The cost function to minimise uses the $p-$norm of a vector to approximate the focusing on the joint which is farthest from its centre, given by:

\begin{equation}
    \Phi \left (q \right) = \left ( \sum_{i=1}^{n} \left | K_{ii}   \frac{q_i-q_{c_i}}{\Delta q_i} \right | ^{p} \right )^{\frac{1}{p}}
\end{equation}
where $q_{c_i}$ is the centre of the joint range $\Delta q_i$ for joint $i$, and $K$ is the matrix for the weights of the joint importance. For the optimised solution to the cost function, $\mathbf{v}$ is solved by:
\begin{equation}
    \mathbf{v}=-\nabla \Phi.
    \label{eq:nabla}
\end{equation}

Equation~(\ref{eq:nabla}) can be optimised for a combined cost function of position and velocity limits. Combining (\ref{eq:dla}) and (\ref{eq:nabla}) of the joint limit costing function, the solution for the required change in joint positions $\mathbf{q_{\theta}}$ is given by:
\begin{equation}
    \mathbf{q_{\theta}} =  \mathbf{Z} \Delta \vec{\mathbf{s}} + \left ( I -  \mathbf{Z}  J\right ) \mathbf{v}.
    \label{eq:jointsolution}
\end{equation}
The weighting for the preference of joint position limit avoidance or joint velocity limit avoidance is able to be customised within OpenSHC.

\begin{algorithm}[t]
\begin{algorithmic}[1]
    \FOR{$h \in [H_{min},H_{max}]$}
        \FOR{$\alpha \in [0,2\pi]$}
            \STATE $endSearch \gets false$,\\
            \STATE $d_{max} \gets 0$, $P_{target} \gets {P_{0}}$\\
            \WHILE{!$endSearch$}
                \STATE $P_{desired} \gets P_{target}$\\
                \STATE $j_{solved} \gets {solveIK}(P_{desired})$\\
                \FOR{$j \in j_{solved}$}
                    \IF{$!{withinJointLimits}(j)$}
                        \STATE $endSearch \gets true$\\
                    \ELSE
                        \STATE $P \gets {solveFK}(j)$\\
                        \STATE ${move}(foot \to P)$\\
                        \IF{$|P-P_{desired}| > \Delta_P$}
                            \STATE $endSearch \gets true$\\
                        \ENDIF
                        \STATE $P_{target} \gets {increment}(P_{target},\alpha, h)$\\
                        \STATE $d_{max} \gets |P_{target} - P_0|$\\ 
                    \ENDIF
                \ENDFOR
            \ENDWHILE
            \STATE $r_{\alpha,h} = d_{max}$
        \ENDFOR
    \ENDFOR
\end{algorithmic}
\caption{Locomotion workspace search}
\label{alg:locomotionworkspace}
\end{algorithm}

\begin{figure*}[!t]
\centering
    \subfigure[FrankenX]{\label{fig:frankenx_ws} \includegraphics[height=6.0cm]{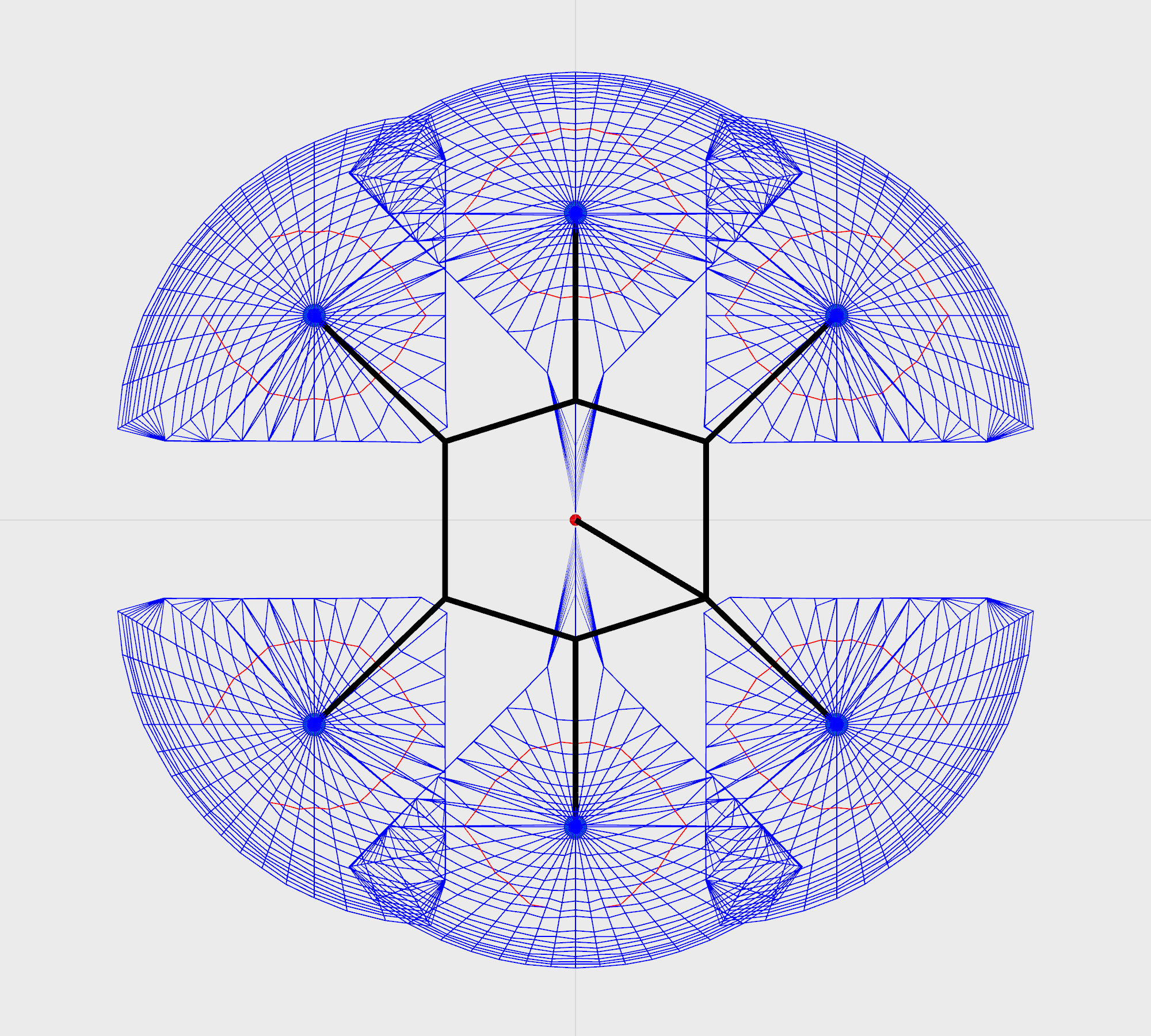}}
    \subfigure[Bullet]{\label{fig:bullet_ws} \includegraphics[height=6.0cm]{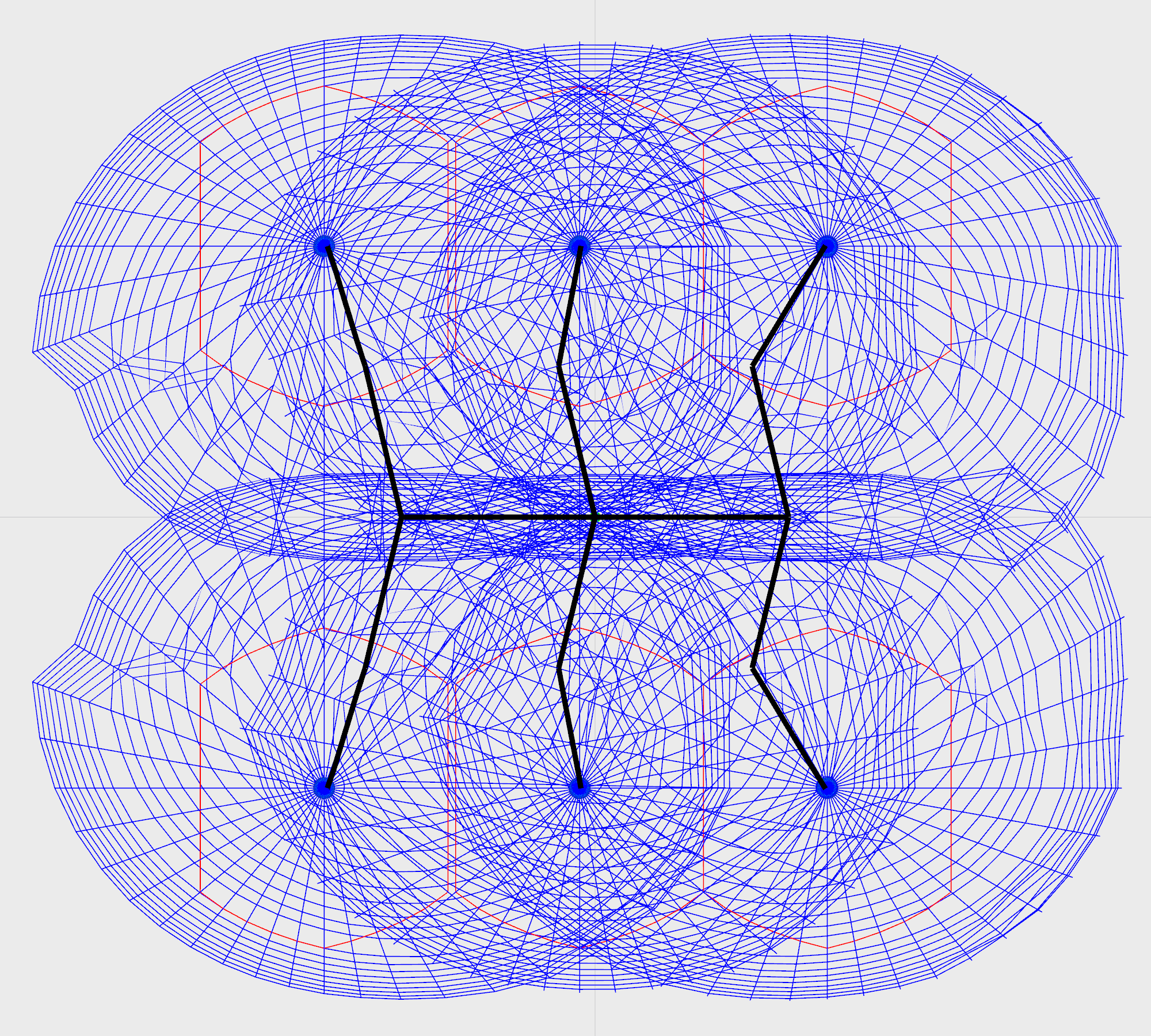}}
    \subfigure[Gizmo]{\label{fig:gizmo_ws} \includegraphics[height=6.0cm]{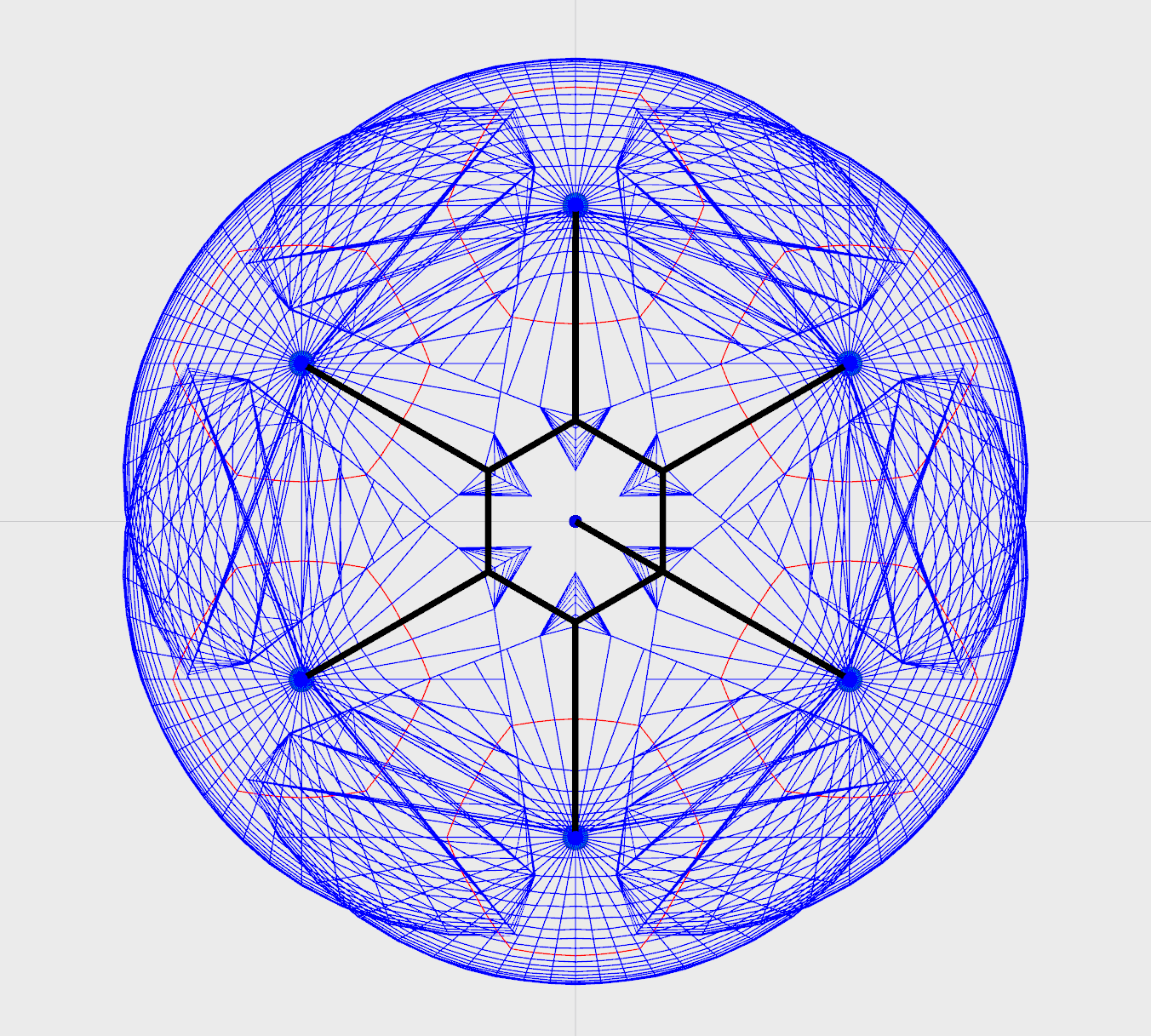}}
    \subfigure[MAX]{\label{fig:max_ws} \includegraphics[height=6.0cm]{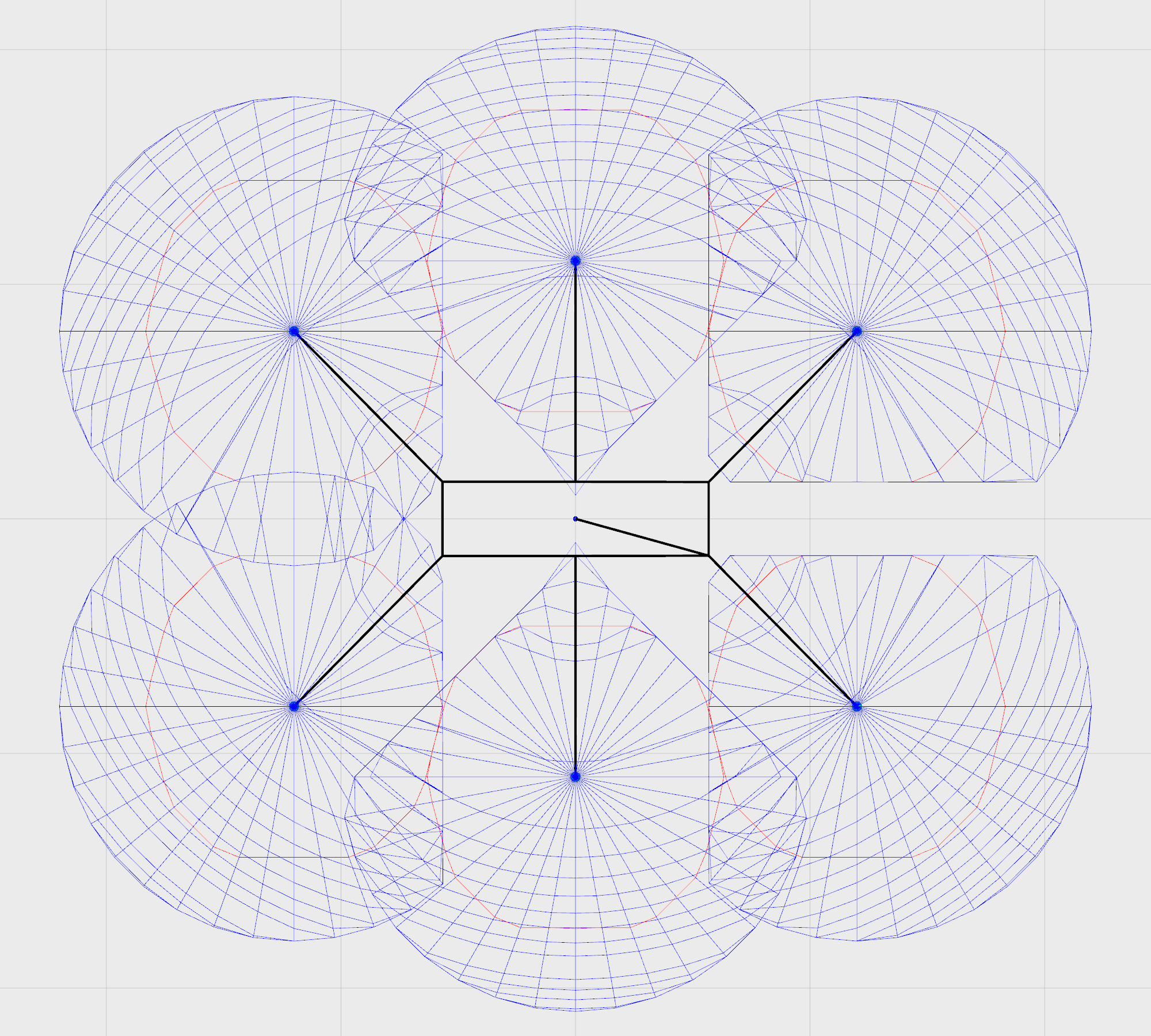}}
    
\caption{{Final result of the locomotion workspace search for four different legged robots visualised in rviz. Top down view: the black outlines represent the robot model, the blue lines show the 3D leg tip workspace limited by morphology and joint limits while the red lines show the restricted planar non-overlapping locomotion sub-space of the workspace referred to as the \textit{walkspace}.}}
\label{fig:footworkspace}
\end{figure*}

\subsection{Leg Workspace}
The locomotion workspace is pre-computed at system initialisation and used to define the workspace available for each leg while walking. {It is centred around the stance foot tip position of each leg and is calculated iteratively through moving the desired foot tip along different directions in the $xy$ plane until the kinematics fail. This process is given in} Algorithm~\ref{alg:locomotionworkspace} {which is executed for each foot in the robot model. Here, $h$ is the search height, incremented by a predefined $\Delta_h$ height step between minimum and maximum search heights $H_{min}$ and $H_{max}$. The symbol $\alpha$ is the search bearing angle, incremented by a predefined $\Delta_{\alpha}$ bearing step. The maximum distance the foot tip travels from its origin position is $d_{max}$. $P_0$, $P_{target}$, $P_{desired}$ and $P$ are the origin tip position, target tip position, desired tip position and actual tip position respectively. The maximum error tolerance between actual and desired tip positions is denoted by $\Delta_{P}$.  The function $solveIK()$ is the inverse kinematics solver producing joint positions for a given foot tip position and $solveFK()$ is the forward kinematics solver producing a foot tip position for a given set of joint angles. $j_{solved}$ is the set of joint angles for a given leg's joints for a desired tip position $P_{desired}$. The workspace radius for each foot at a given search bearing $\alpha$ and height $h$ is $r_{\alpha,h}$.}

{The result is a volumetric workspace, with the different $z$-axis height slices forming a polyhedron for each leg. The top-down view of these polyhedra for four different robots is illustrated in} Fig. \ref{fig:footworkspace}.
{A single restricted workspace polygon at the desired body height is called the \textit{walkspace}.
The walkspace is constrained to be symmetrical, with the minimum walkspace from all the legs selected so that the polygon is the same for each leg. The combination of non-overlapping workspaces and JLA are used to prevent self-collisions without a dedicated collision check module. 
The walkspace is also used to calculate and limit the stride length for a given desired input body velocity. The desired body velocity is passed to the robot for a given step frequency, with the corresponding stride length calculated to achieve the desired body velocity. The maximum linear body velocity is thus limited by the walkspace of the legs.} The body velocity calculated by OpenSHC is given by:
\begin{equation}
    v_{body} = \frac{ l_{s} \times  f_{s}}{\beta}
    \label{eq:velocity}
\end{equation}
where $\beta$ is the duty factor defined as $\beta = T_{stance} / T_{stride}$, the time in stance phase, and the total time of stance and swing phase for the stride time; $l_s$ the stride length; and $f_s$ the step frequency.
Note that both $l_{s}$ and $f_{s}$ are not fully independent, as $f_s$ is limited by the maximum joint velocity, which is also affected by the required distance to move from the stride length.
In reality, the actual velocity is lower due to slippage and other disturbances. However, this is not accounted for in OpenSHC. Therefore, for accurate robot odometry, an external tracking solution is required. More examples of leg workspaces are presented in Section~\ref{sec:legkinematicworkspace} and Fig.~\ref{fig:workspaces}.


\section{System Architecture}
\label{sec:systemarchitecture}

\begin{figure*}[!t]
\centering
    \subfigure[]{\label{fig:rviz_weaver} \includegraphics[height=4.5cm]{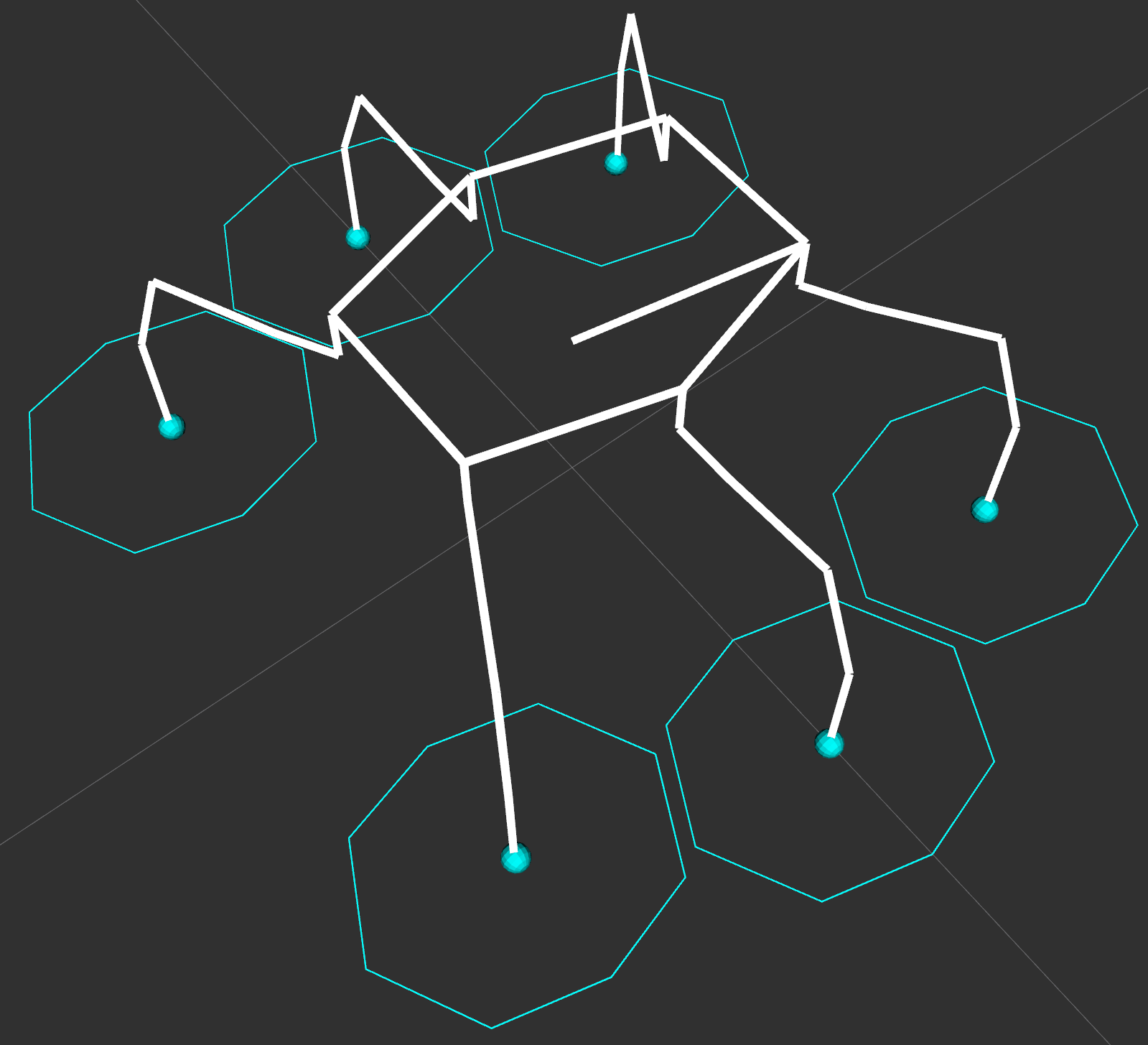}}
    \vspace{-0.1cm}
    \subfigure[]{\label{fig:gazebo_weaver} \includegraphics[height=4.5cm]{gazebo_weaver.png}}
    \vspace{-0.1cm}
    \subfigure[]{\label{fig:real_weaver} \includegraphics[height=4.5cm]{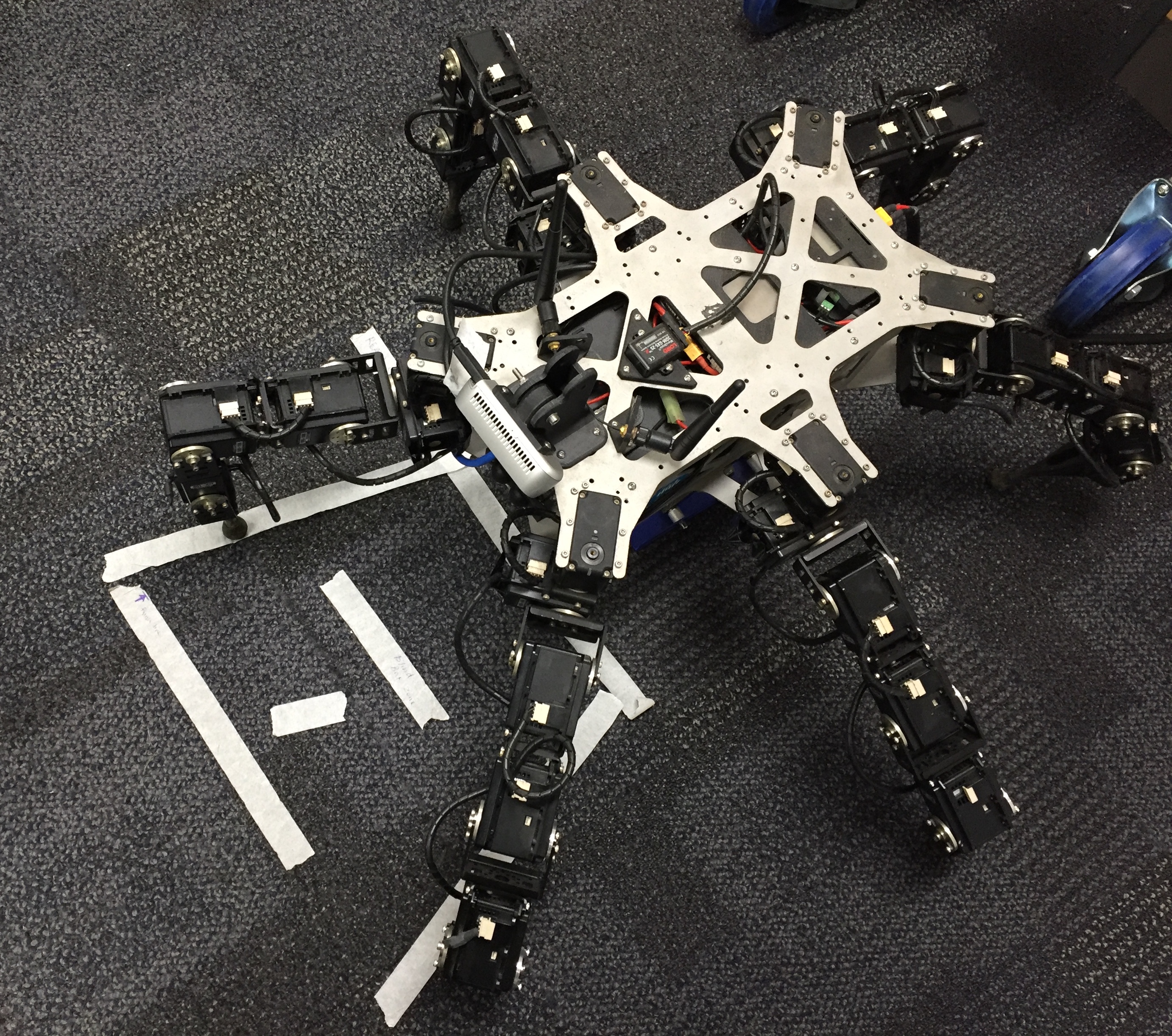}}
    \vspace{-0.1cm}
\caption{The hexapod Weaver in (a) rviz kinematic model, (b) Gazebo simulation and (c) hardware.}
\label{fig:rvizgazeboreal}
\end{figure*}

{The high-level controller consists of multiple modules, and is wrapped as a C++ node within the Robot Operating System (ROS)} \cite{quigley_2009} {framework.} The utilisation of rostopics and the ROS parameter server allows for a common interface to work with various inputs (for control and sensors) and outputs (position control of joints), and easy dynamic parameter customisation. With the integration of Gazebo \cite{koenig_gazebo} with ROS, robot algorithms can be tested in simulation before being deployed onto hardware. Fig.~\ref{fig:rvizgazeboreal} shows the robot Weaver as a pure kinematic model in rviz, a simulation model in Gazebo and the real robot hardware.

\begin{figure}[b!]
    \centering
    \includegraphics[width=85mm]{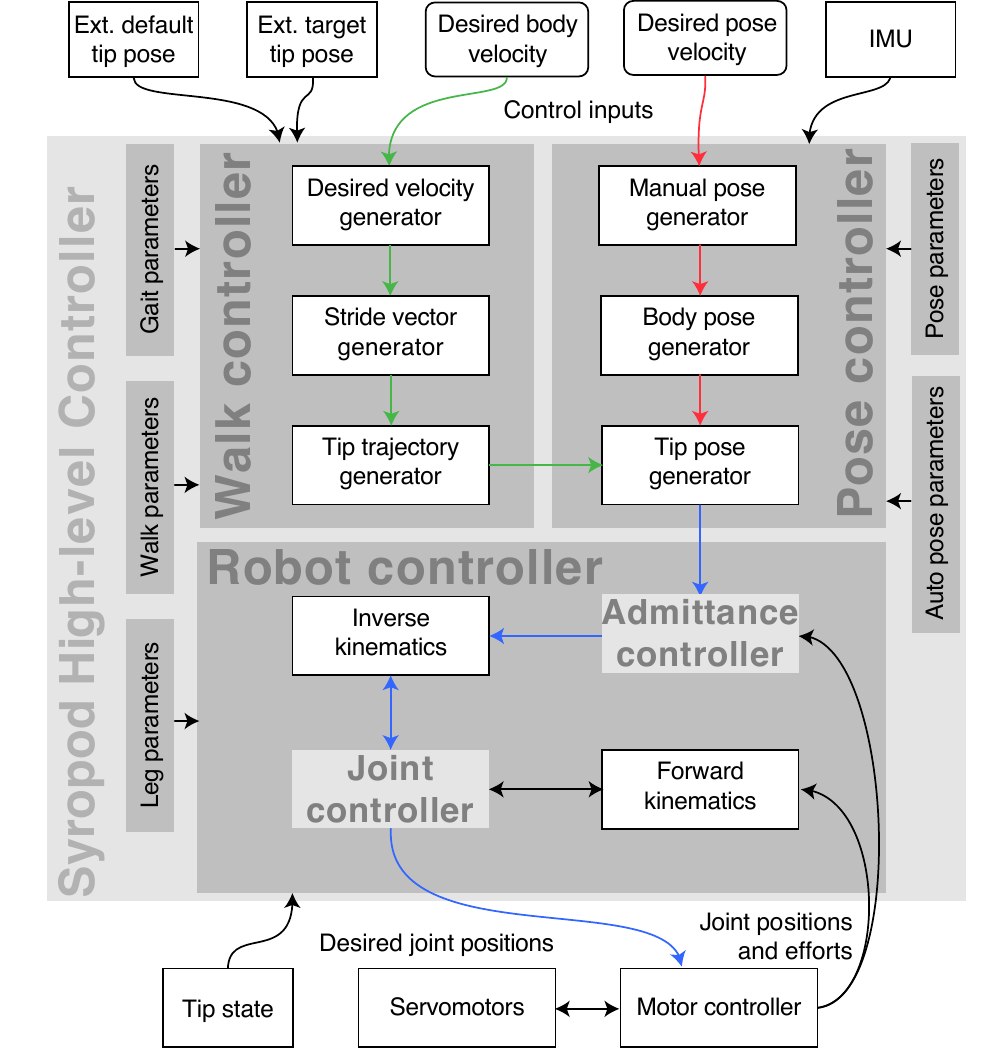}
    \caption{OpenSHC's hierarchical control architecture shown in a simplified view containing typically used components only. The coloured arrows show the flow of information in each controller during typical use. The green, red and blue lines show control flow through the walk controller, the pose controller and the robot controller respectively. 
    }
\label{fig:control_arch}
\end{figure}

{The modules of OpenSHC can be customised via configuration parameters for each robot platform, with additional functionality modules configurable to be enabled or disabled depending on the scenario. For example, for inclined, uneven terrain, the additional functionality of the different pose generators} (Section~\ref{sec:posecontroller}) and admittance controller (Section~\ref{sec:legcontroller}) can be enabled to increase stability.
 Fig.~\ref{fig:control_arch} {shows the simplified structure of OpenSHC with its core functionality. The coloured lines show control flow in typical use. The subsequent figures breakout each of the three main controllers in to greater detail. These are the walk controller in} Fig.~\ref{fig:walk_control_arch}, {the pose controller in} Fig.~\ref{fig:pose_control_arch} {and the robot controller in} Fig.~\ref{fig:robot_control_arch}.
 
 {The control inputs received by OpenSHC passes the desired body velocity and pose velocity to the walk controller and pose controller whose control flows are shown in green and red respectively. The tip pose generator combines the tip trajectory with the body pose and passes it to the robot controller.}

 This takes the resultant tip pose and calculates the desired joint angles. The joint controller calls the IK and FK systems and performs a final check to constrain the output joint positions to safe limits. The output joint states of OpenSHC are either sent to the motor controller which controls the servomotors, or to the control interface of a simulation engine such as Gazebo. Typical control flow through the robot controller is shown in blue.

\subsection{Control Inputs}
\label{sec:controlinput}
OpenSHC receives desired body velocities (forward, lateral and angular) and body posing velocities (6 DOF) via an external source. The source can be an operator via a gamepad, tablet, computer, or from an autonomous navigation stack. The interface takes in a linear velocity vector along the $x,y,z$ axes and an angular velocity vector about the $x,y,z$ axes, using the ROS \textit{geometry\_msgs::Twist} message as transport. 
{
For the walk controller, the system uses the linear $x$ and $y$ velocities and angular $z$ velocity, with linear $z$ velocity and angular velocities about $x$ and $y$ being ignored.
For the pose controller, all six velocities are used to pose the robot's body.}

{The user is able to select which external source to use, allowing auto navigation or cruise control mode, along with manual control. Auto navigation requires commands from an autonomous navigation stack that provides obstacles avoidance functionality, while cruise control sets the robot's velocity constant in the desired velocity generator within the walk controller.}

\subsection{Walk Controller}
\label{sec:walkcontroller}
\begin{figure}[b!]
    \centering
    \includegraphics[width=85mm]{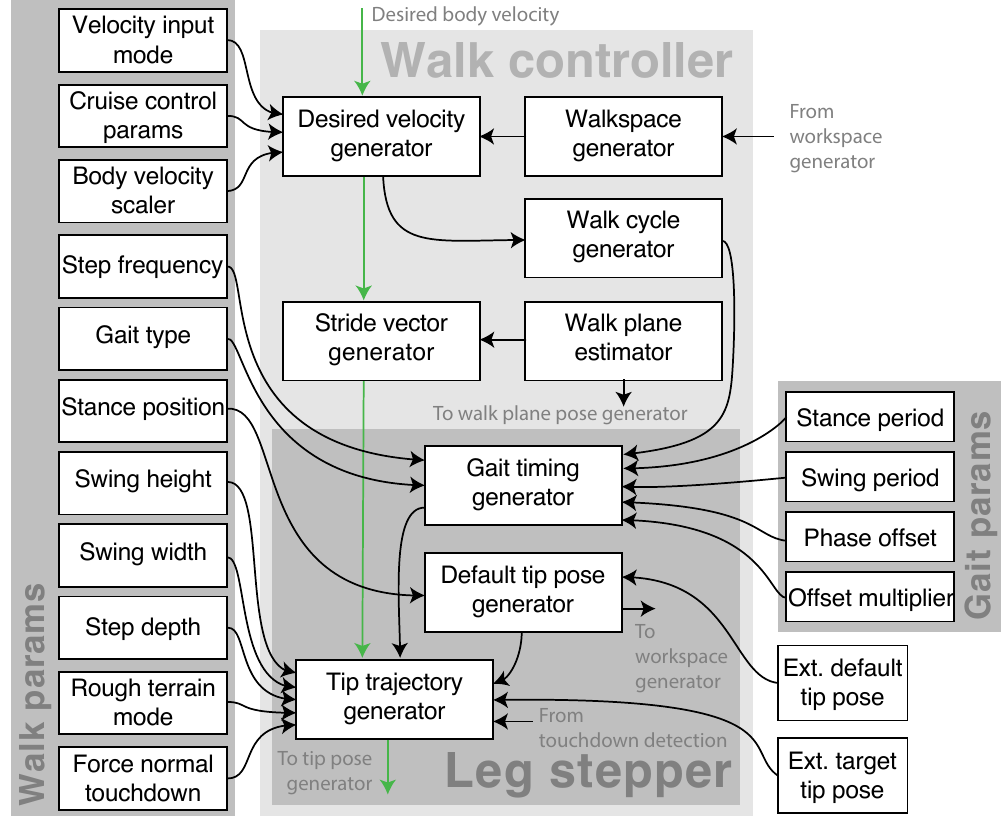}
    \caption{Detailed diagram of the walk controller with typical control flow shown in green.}
\label{fig:walk_control_arch}
\end{figure}

{The walk controller} (Fig.~\ref{fig:walk_control_arch}) {generates the required foot tip trajectories and timings from the desired body velocity. Within the walk controller, the gait configuration parameters are converted into timings for each leg in the gait timing generator} (Section~\ref{sec:gaitengine}) {which are then transformed to foot tip trajectories in the tip trajectory generator} (Section~\ref{sec:trajectoryengine}). {The desired velocity generator fits the body velocity to the walkspace while the stride vector generator calculates the required stride length using} (\ref{eq:velocity}) {to achieve the desired velocity. The following subsections describe the gait timing generator and the tip trajectory generator, two major components of the walk controller, in detail.}

\subsubsection{Gait Timing Generator}
\label{sec:gaitengine}
\begin{figure}[t!]
    \centering
    \includegraphics[width=75mm]{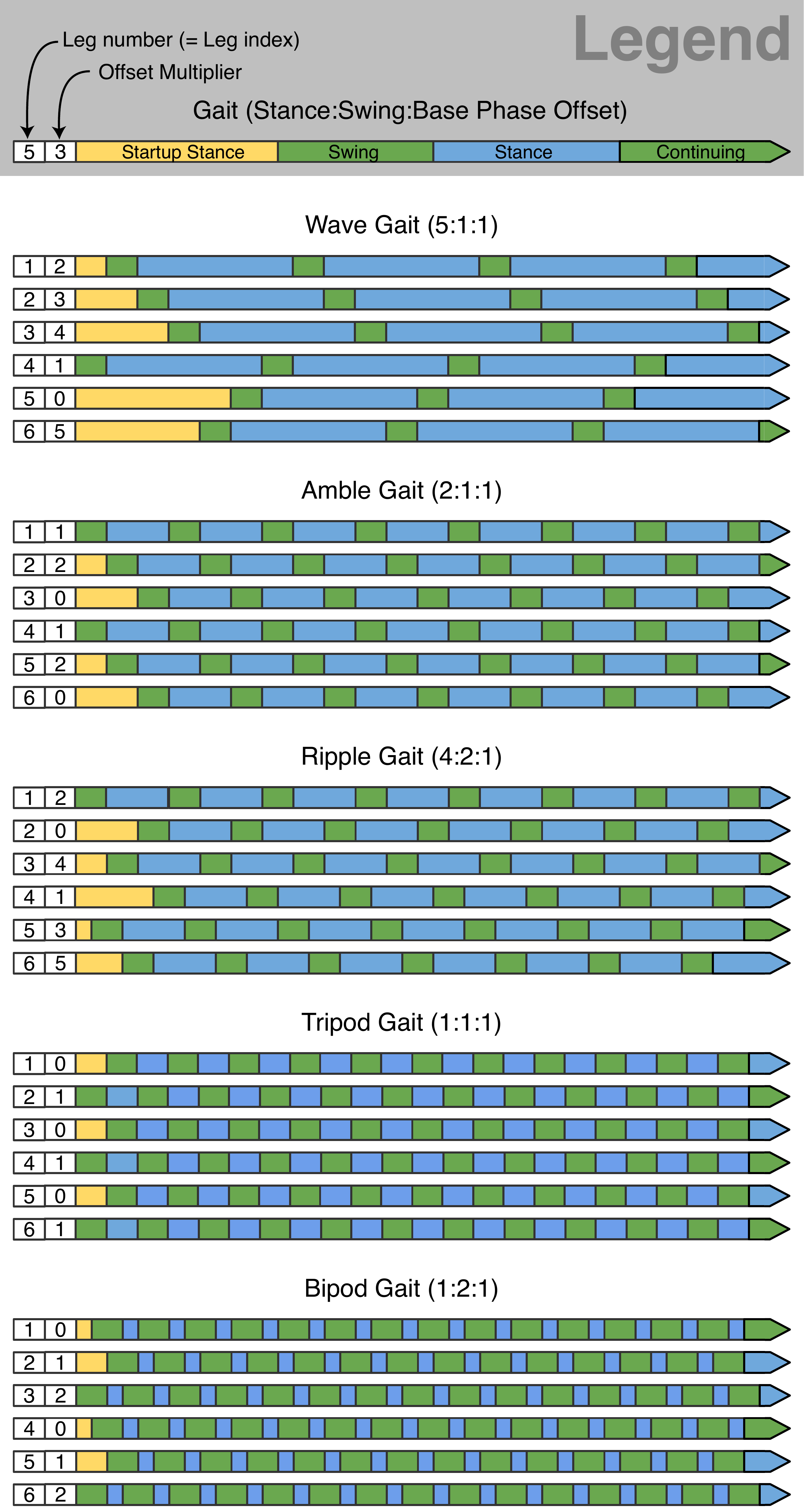}
    \caption{Timing for predefined gaits for a hexapod robot.}
\label{fig:gaittiming}
\end{figure}

Working along with the stride vector generator to achieve the desired velocity, the gait timing generator takes in predefined parameters defining the different gaits; such as wave, amble, ripple, tripod and the dynamic bipod gaits \cite{kottege2015energetics,Ramdya2017}. The parameters define the ratio between the stance and swing periods of each gait cycle and the phase offset between each leg executing the gait cycle, as shown in Fig.~\ref{fig:gaittiming}. The timing of these gait cycles and the desired state of the leg is sent to the tip trajectory generator which generates either a swing or stance trajectory based on the current state of the leg in the gait cycle.

\subsubsection{Tip Trajectory Generator}
\label{sec:trajectoryengine}
The tip trajectory generator calculates the foot tip trajectories for the specified gait pattern and stride vector. It consists of three $4^{\text{th}}$ order B\'{e}zier curves to generate foot positions and velocities over the time period of the step cycle. Two B\'{e}zier curves control the foot trajectory across the primary (first half) and secondary (second half) swing period, and the third B\'{e}zier curve controls the stance period. The stance B\'{e}zier curve generates foot velocities rather than positions using the derivative of the position B\'{e}zier curves used in the swing period. This is to ensure that on-ground foot velocities are guaranteed to match those required to achieve the desired body velocity. Each B\'{e}zier curve is defined by 5 control points and the following equations:
\begin{equation}
    B(t) = s^4P_0 + 4ts^3P_1+6s^2t^2P_2+4st^3P_3+t^4P_4
    \label{eq:beziercurve}
\end{equation}
where $s = 1-t$ and $t \in [0,1]$. The derivative of (\ref{eq:beziercurve}) is given by:
\begin{multline}
    {B}'(t) = 4s^3 (P_1-P_0)+12s^2t(P_2-P_1) \\ +12t^2s(P_3-P_2)+4t^3(P_4-P_3).
    \end{multline}

\begin{figure}[t!]
    \centering
    \includegraphics[width=75mm]{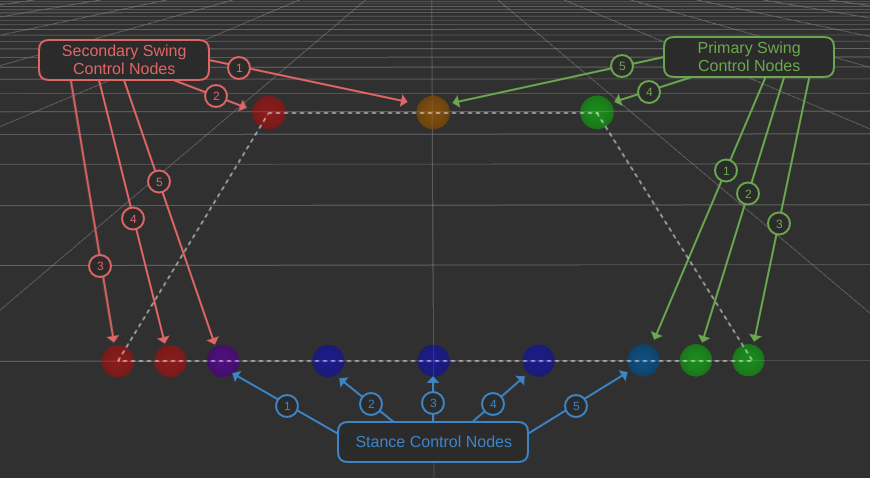}
    \caption{The control points for the three B\'{e}zier curves that form the foot tip trajectory.}
\label{fig:trajectory_controlpoints}
\end{figure}

The control points are placed to maximise smoothness throughout the step cycle whilst ensuring desired swing characteristics, such as step clearance and liftoff/touchdown placement, as well as required ground velocity during stance. Position continuity is ensured between step cycle periods, as shown in Fig.~\ref{fig:trajectory_controlpoints}, by setting the `end' control points for each B\'{e}zier curve at the same location. Similarly, control point placement strategies are employed for the remaining control points to ensure velocity and acceleration continuity, producing trajectories which are at least $C^1$ smooth and preferably $C^2$ smooth within the swing characteristic constraints. The desired characteristics of the swing and stance periods have precedence over trajectory smoothness. OpenSHC ensures that the foot trajectories will always adhere to three requirements: 
\begin{itemize}
  \item The foot trajectory during swing is position controlled to ensure it achieves the defined liftoff and touchdown positions at the start and end respectively of the primary and secondary swing B\'{e}zier curves.
  \item The foot trajectory during swing is position controlled to ensure it achieves the highest point at the desired step clearance directly above a defined `default' foot position, i.e. the position of the foot with zero body velocity.
  \item The foot trajectory during stance is velocity controlled to ensure the required foot velocity is attained as per the requirements of the desired body velocity.
\end{itemize} 

\begin{figure}[t!]
    \centering
    \includegraphics[width=75mm]{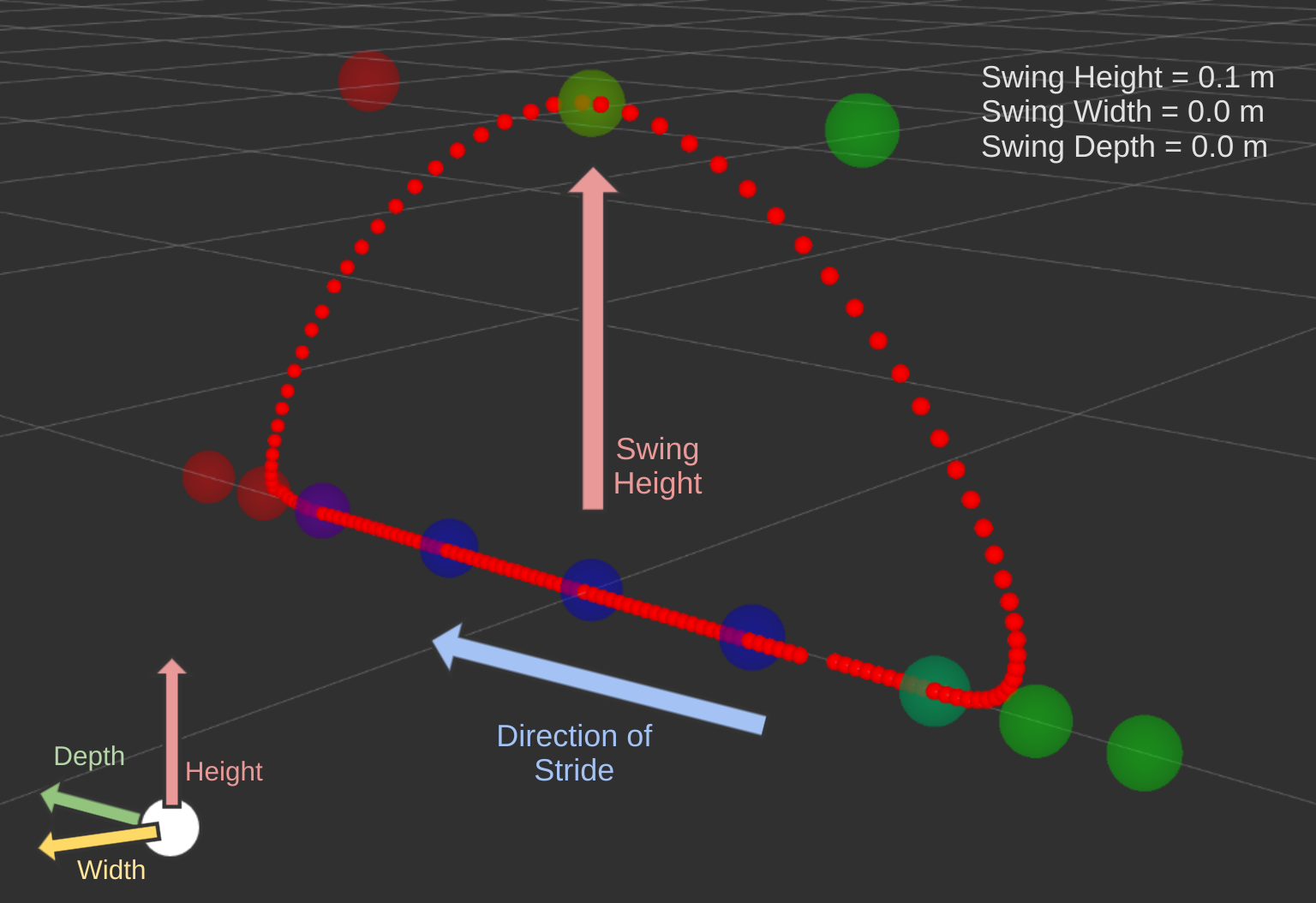}
    \caption{The foot tip trajectory as defined by the three B\'{e}zier curves and associated control points, ensuring desired stride length, step clearance, step width and swing depth.}
\label{fig:trajectory}
\end{figure}

The resultant foot trajectory from the control points is illustrated in Fig.~\ref{fig:trajectory}.

\subsubsection{Manual Leg Manipulation}
\label{sec:legipulation}
{A`freegait' module, within the walk controller but not shown in} Fig.~\ref{fig:walk_control_arch} {due to being decoupled from the typical control flow, allows for leg tip poses to be controlled outside of the cyclic gait. The desired tip pose of any leg can be controlled by either specifying the velocity of the leg tip or the actual pose of the leg tip in reference to the leg frame for basic manual leg manipulation tasks (legipulation). The freegait tip pose overrides the tip trajectory generator output to the tip pose generator. Achieving the desired tip pose is not guaranteed, subject to IK being able to solve for valid joint angles for the DOFs available.}

\subsection{Pose Controller}
\label{sec:posecontroller}
\begin{figure}[t!]
    \centering
    \includegraphics[width=85mm]{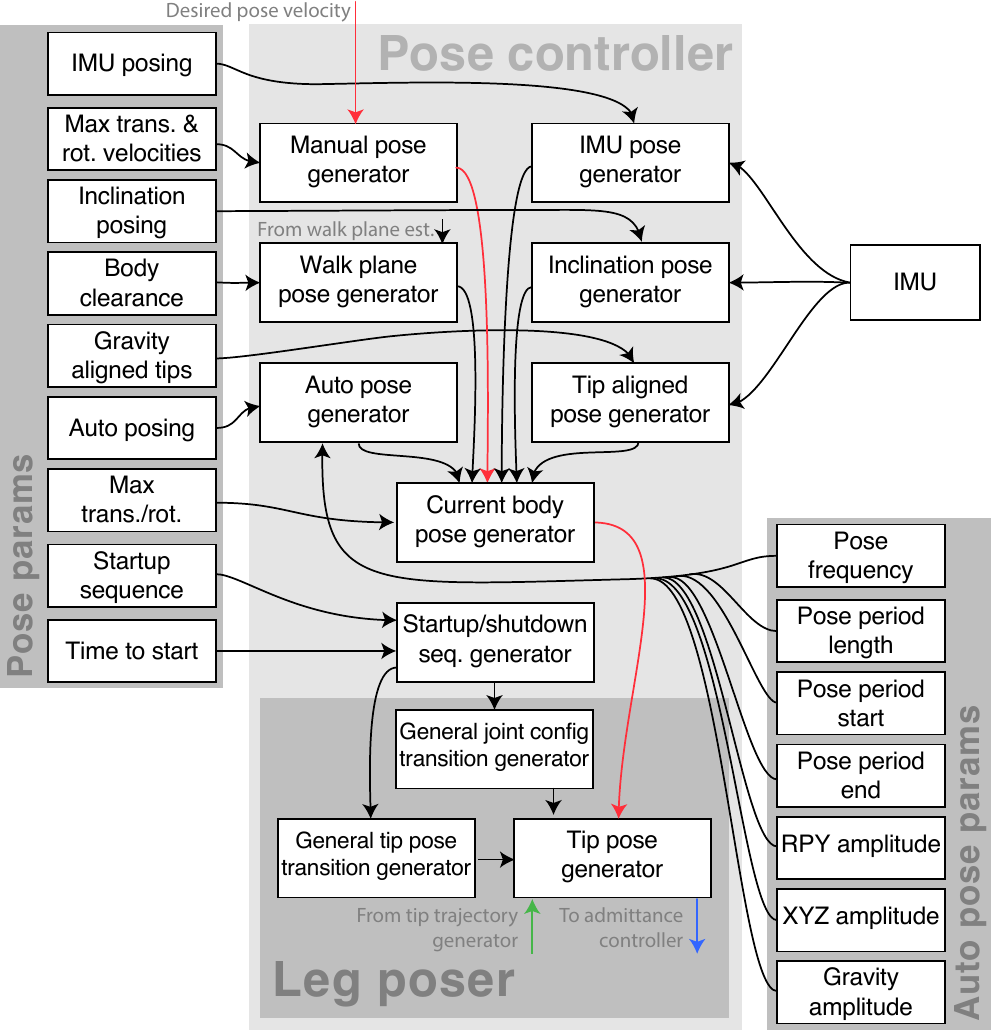}
    \caption{Detailed diagram of the pose controller showing typical control flow in red.}
\label{fig:pose_control_arch}
\end{figure}

The pose controller (Fig.~\ref{fig:pose_control_arch}) generates the required tip pose to achieve the input manual body pose and sensor based body pose for safe operation. Changing the body pose from the default position and orientation allows the robot to shift its centre of mass for increased stability on rough terrain. {Additionally, sensors mounted onto the robot's body with limited field of view can be oriented towards an area of interest by changing the body pose.}
{The manual pose generator receives the desired body posing velocities from the control input  and alters the robot's body pose within the maximum limits.}

{The following subsections give details of five main components of the pose controller: IMU pose generator, inclination pose generator, current body pose generator, startup/shutdown sequence generator and the tip pose generator.
}

\subsubsection{IMU Pose Generator}
\label{sec:IMUpose}
The IMU pose generator provides feedback to correct the robot's body position. The IMU pose generator makes use of the inertial measurement unit (IMU) to ensure the robot body does not tend towards an unstable position. As the body experiences changes in roll and pitch, it is posed in opposing roll and pitch axes to correctly align the robot frame $z$-axis parallel to the direction of the gravity vector to keep the body level. This posing controller uses a PID loop to ensure the posing is stable and remains within predefined limits.

\subsubsection{Inclination Pose Generator}
\label{sec:inclination}
 During locomotion in heavily inclined terrain, if the body is oriented such that correction from posing in roll/pitch cannot stabilise it, then the inclination pose generator poses the body laterally in the $xy$ plane of the robot frame according to the IMU orientation input. The goal of the system is to centre the vertically projected robot centre of mass into the centre of the support polygon of the robot's load bearing legs.

\subsubsection{Current Body Pose Generator}
\label{sec:bodypose}
{The current body pose generator adds together the output of several separate body posing systems whilst ensuring legs do not exceed the maximum translation and rotation limits of the body.} 
In addition to the two main body posing sub-systems in Sections~\ref{sec:IMUpose} and \ref{sec:inclination}, the body pose generator takes input from: {the walk plane pose generator which aligns the body to a \textit{walkplane} estimate; the auto pose generator which is an automatic user defined cyclical body pose generator; and the tip aligned pose generator which shifts the body to allow 3\,DOF legs to achieve desired tip orientations upon touchdown.
Each body posing sub-system transforms the current body pose generator output $P_{body}$ as summarised by:}

\begin{equation}
P_{body} = H_{ali} H_{AI} H_{inc} H_{man} P_{walk}
\end{equation}
{where $H_{man}$, $H_{inc}$ and $H_{ali}$ are the transformation matrices calculated by manual pose, inclination pose and tip align pose respectively. $H_{AI}$ is either IMU pose or auto pose, as only one can be active at a time without interference. $P_{walk}$ is the pose parallel to the walkplane, which is the plane of best fit from the foot tip contact points. The reference frame of $P_{walk}$ is $P_{default}$, the default pose given by zero translation and identity orientation, except for the body clearance. 
The output poses of sub-systems that have similar functionality and incompatible with each other (such as IMU pose or auto pose) are removed from the output pose to ensure the system is not overcompensating for disturbances.} 

\subsubsection{Startup/Shutdown Sequence Generator}
\label{sec:sequencegenerator}
The sequence generator turns a list of joint angles for each joint to achieve sequentially, to moving the robot joints to the desired angles. This provides smooth motions from any initial leg position to the leg positions for stance. The sequence generator also allows for the robot to move to a packed state where the legs are tucked in for easy transport. Intermediate steps allows for overlapping packed legs to be unpacked safely without leg collisions. 

\subsubsection{Tip Pose Generator}
\label{sec:tippose}
{The tip pose generator combines the desired body pose output} of the body pose generator and the desired tip pose output of the tip trajectory generator to create the resultant tip pose for each leg of the robot.
{Each leg tip pose $P_{leg}$ is referenced from the default body pose frame $P_{default}$, the same reference frame used for $P_{walk}$. Thus, the inverse transform is used to transform the leg tip pose $P^{default}_{leg}$ to $P^{body}_{leg}$.}

The result of each leg's tip pose generator enables the robot to enact desired foot tip trajectories for desired walking characteristics whilst simultaneously posing the body as desired. This module also ensures tip poses do not exceed the given workspaces for each leg.

\begin{figure}[t!]
    \centering
    \includegraphics[width=85mm]{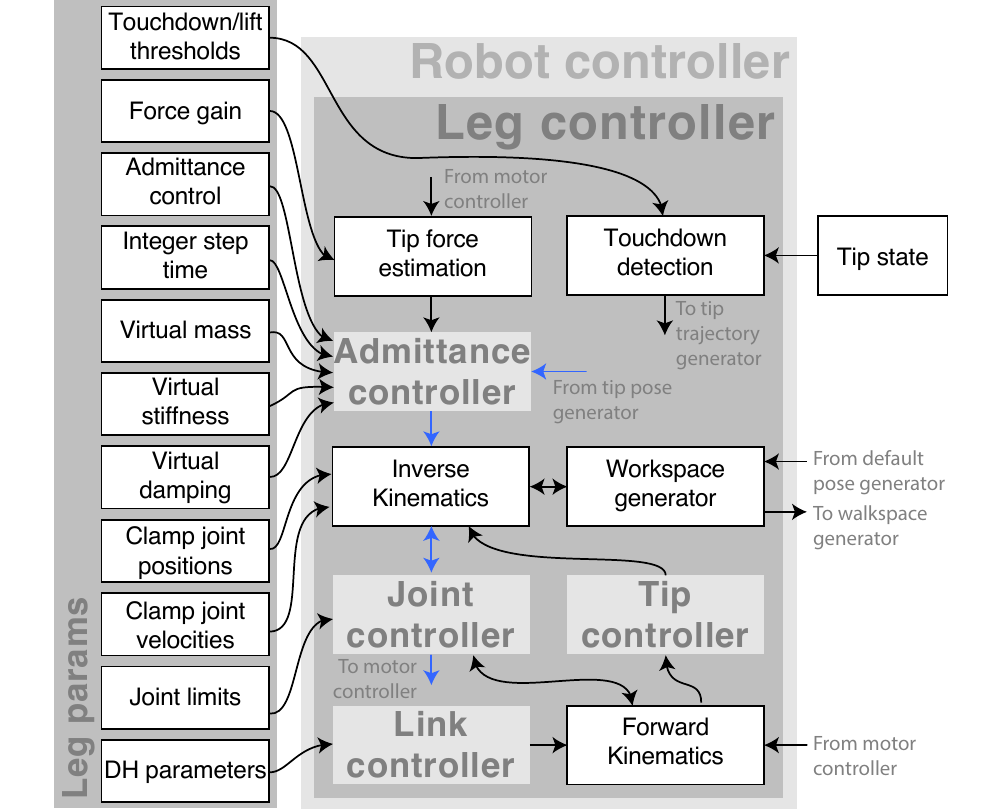}
    \caption{Detailed diagram of the robot controller showing typical control flow in blue.}
\label{fig:robot_control_arch}
\end{figure}

\subsection{Robot Controller}
\label{sec:legcontroller}
{The robot controller} (Fig.~\ref{fig:robot_control_arch})  {takes the desired poses and timings of the foot tip and treats it independently for each leg via the leg controllers. It is able to detect foot tip touchdown to stop the leg motors over-torquing and to model the walk plane of the robot. the following subsections describe two of the main controllers within the leg controllers; admittance controller and joint controller.}

\subsubsection{Admittance Controller}
\label{sec:admittance}
\begin{figure}[t!]
    \centering
    \includegraphics[width=75mm]{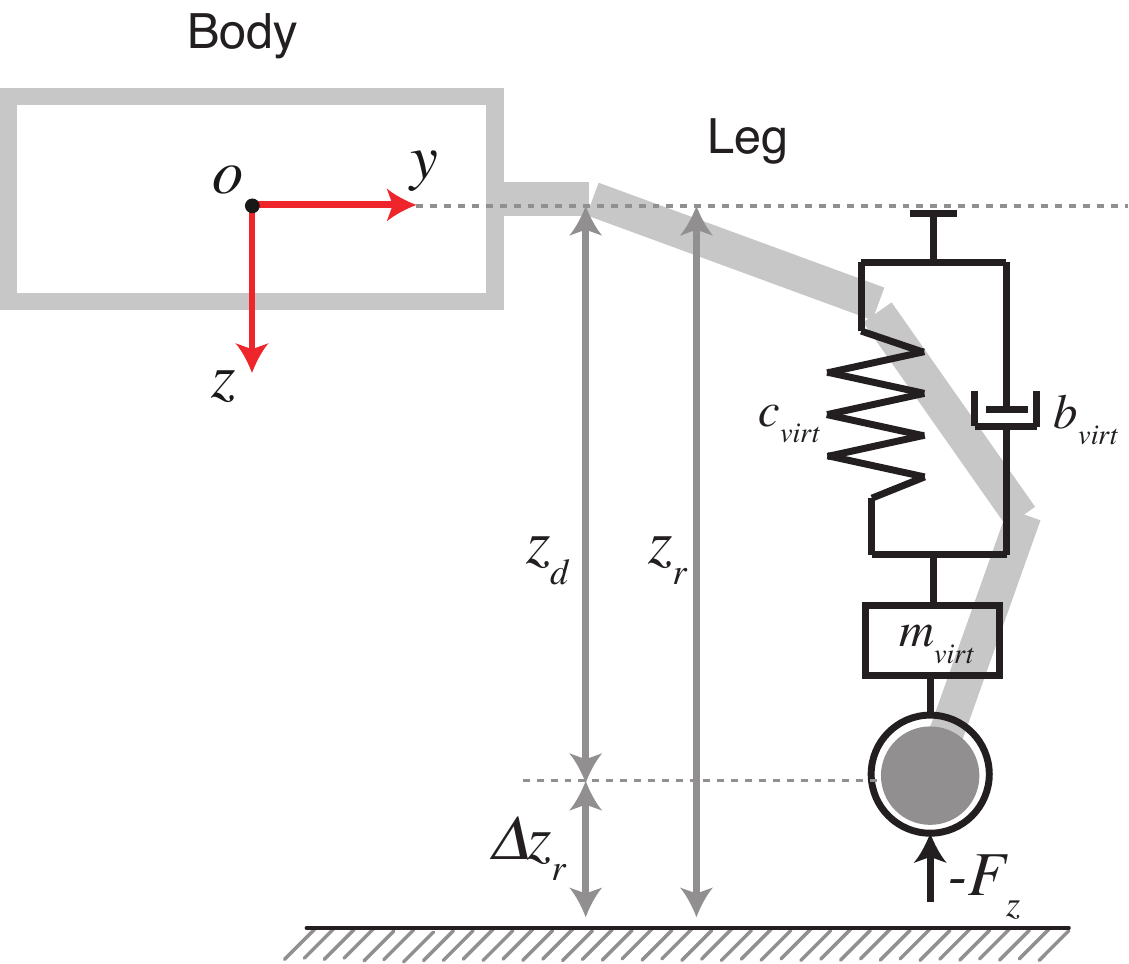}
    \caption{Admittance control modelled as a mass spring damper mechanical system.}
\label{fig:admittance}
\end{figure}

The admittance-based leg controller is used for rough terrain locomotion and is based on the research in \cite{Bjelonic2016}. A virtual elastic element is added to the legs to compensate for non-planar foot placements. Joint torques are mapped to the force at the foot tip using a Jacobian $J_{e}$ as follows:
\begin{equation} \label{eq:forceTransmission}
\begin{bmatrix}
F_{e}^{2}(t) \\
M_{e}^{2}(t)
\end{bmatrix} =
(J_{e}(q_{1}, q_{2}, q_{3}, q_{4}, q_{5})^T)^{-1} \cdot M_q
\end{equation}
where $M_q=[M_1~M_2~M_3~M_4~M_5]^T$ is the vector of joint torques and $q_{1}, q_{2}, q_{3}, q_{4}, q_{5}$ are the five joint angles corresponding to joints coxa$_\textrm{yaw}$, coxa$_\textrm{roll}$ femur, tibia and tarsus. $F_{e}^{2}=[F_x~F_y~F_z]^T$ and $M_{e}^{2}=[M_x~M_y~M_z]^T$ represents the 3D force  and torque vectors at the foot tip. The admittance controller takes the force at the foot tip as an input and outputs a displacement along the $z$ axis. As shown in Fig.~\ref{fig:admittance}, a virtual mass $m_{virt}$, virtual stiffness $c_{virt}$ and virtual damping element $b_{virt}$ defines the dynamic behaviour along the $z$ axis. This second order system is represented by:
\begin{equation} \label{eq:impedanceAdv}
-F_{z}=m_{virt}\ddot{\Delta z_{r}}+b_{virt} \dot{\Delta z_{r}}+c_{virt} \Delta z_{r}.
\end{equation}

The exerted force and the foot position is adapted by this virtual second order mechanical system by reacting to measured foot force. This adapted foot position is $z_{d}=z_{r}-\Delta z_{r}$. The displacements $\Delta x_{r}$ and $\Delta y_{r}$ are set to zero in order to restrict motion in the $x$ and $y$ directions during foot contact. 

\subsubsection{Joint Controller}
\label{sec:jointcontroller}
The joint controller handles interaction with the forward and inverse kinematics systems for every joint in each leg to achieve a desired foot tip pose. It provides leg state and model information for the associated joints for the FK and IK systems to solve (\ref{eq:FKtransform}) and (\ref{eq:jointsolution}) respectively. It also restricts any joints to ensure position and velocity limits are not exceeded. The controller acts with the FK and IK systems to produce joint commands which, once sent to the motors, will execute the desired pose of the associated foot tip as close as possible whilst adhering to joint limits.

{To convert the desired joint states to movement of the motors, motor interface packages are required. The integration with ROS allows for modularity with different servo motors in the robot hardware. For simulation, a Gazebo model of the robot is controlled through the output of OpenSHC using the ROS joint\_state\_controller} \cite{ros_control}.


\section{Case Study: Optimal Legged Locomotion}
\label{sec:casestudy}

\begin{figure}[t!]
    \centering
    \includegraphics[width=85mm]{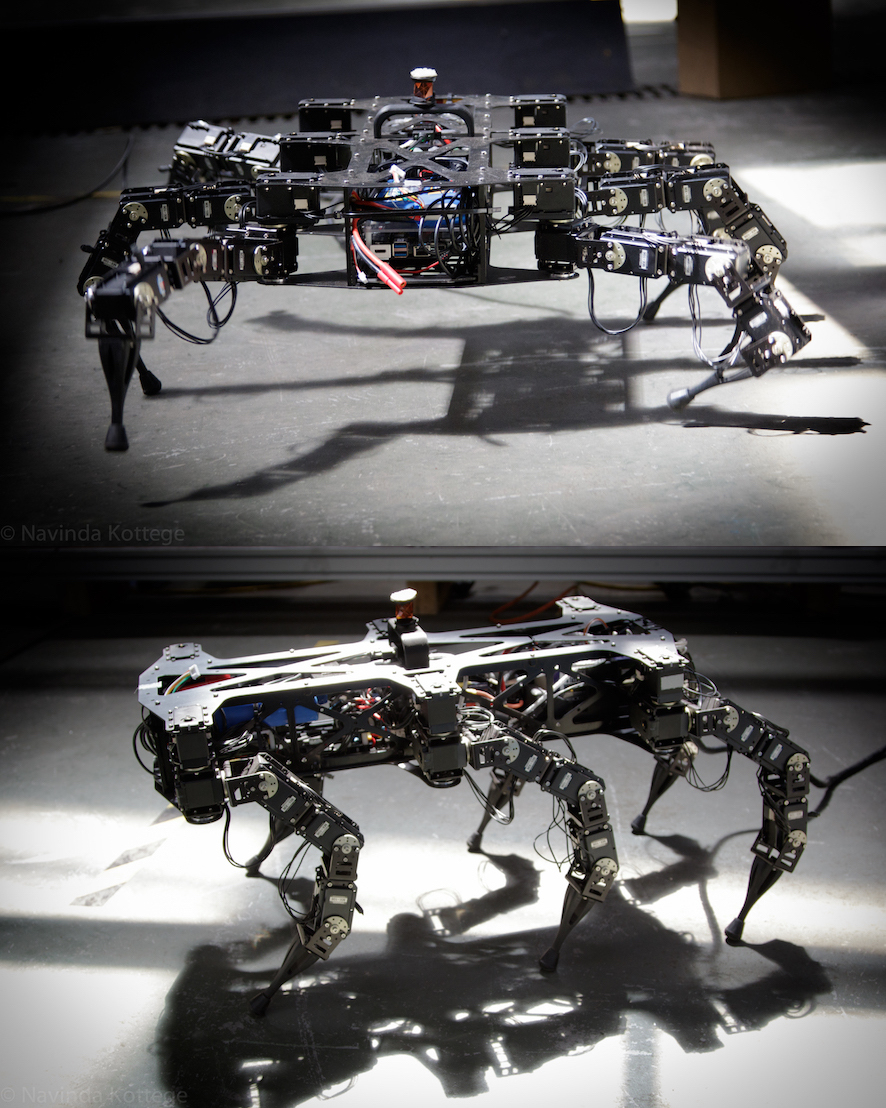}
    \caption{Hexapod robot Bullet walking in insectoid (top) and mammalian (bottom) configurations.}
\label{fig:bullet}
\end{figure}

The versatility of OpenSHC enables rapid customisation of the locomotion parameters to enable different hardware configurations to move. The underlying kinematics and trajectory algorithms allow various parameters to be optimised. Coupled with 3D printing for rapid hardware iterations, OpenSHC is used for hardware optimisation of the locomotion parameter space for a hexapod robot named Bullet, shown in Fig.~\ref{fig:bullet}. 

 \begin{figure*}[!t]
    \centering
    \includegraphics[width=140mm]{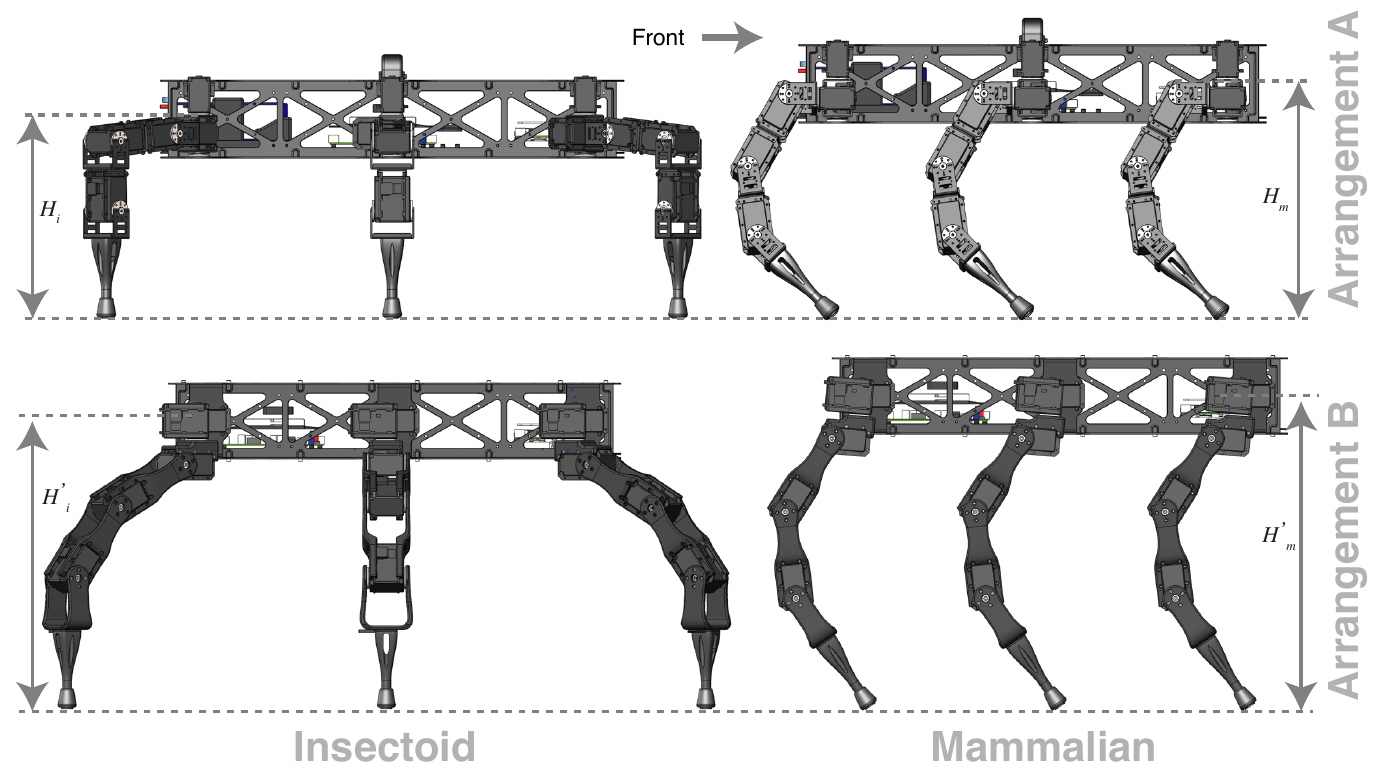}
    \caption{Bullet in insectoid and mammalian configuration in kinematic arrangements A and B. 
    }
    \label{fig:stancesAB}
\end{figure*}

The mammalian and insectoid configurations are favoured in nature for different animals with different number of legs. Predominately, four legged animals have a mammalian configuration while six legged animals have an insectoid (arachnid/sprawling-type) configuration. This is reflected in robotic research as well; mammalian quadruped robots such as ANYbotics' ANYmal \cite{Hutter2016}, Boston Dynamics' Spot (successor of BigDog \cite{Raibert2008}),  IIT's HyQ \cite{semini_design_2017} and MIT's Cheetah~3 \cite{bledt_mit_2018} have sophisticated controllers capable of dynamic locomotion that are fast and agile. Insectoid hexapod robots, such as MAX \cite{Elfes2017}, LAURON V \cite{Roennau2014}, DLR-Crawler \cite{gorner_2018} and SILO6 \cite{Gonzales2007} have been specifically designed for the rough terrain found in disaster zones, demining areas and complex environments. For effective locomotion, the mammalian configuration uses a roll-pitch-pitch configuration for its leg joints. The pitch joints propel the body forwards, while the roll joints are used for stability and to aid steering. For insectoid, a yaw-pitch-pitch configuration is used, with the yaw joints mainly used to propel the body and steer. The pitch joints are required to bear the body weight, even when stationary, consuming more energy compared to mammalian configuration. 

Research in unique locomotion principles have resulted in designs of insectoid quadrupeds and mammalian hexapods. 
The TITAN XIII \cite{Kitano2016}, aimed to use the wider range of leg motion of an insectoid configuration on a quadruped to achieve energy efficient dynamic walking. MRWALLSPECT-III \cite{Kang2003}, a climbing robot also used this design to allow for a greater workspace with a low mass. SILO6, a hexapod robot, has been analysed for energy efficient configurations in \cite{Sanz2012}. Theoretical and experimental analysis conducted on flat terrain showed that the mammalian configuration in a hexapod produced a lower power consumption overall, compared with insectoid. However, SILO6 could not switch between insectoid and mammalian configurations on-the-fly and required manual intervention to mechanically change the leg mounting for this to be possible. Theoretical analysis of mammalian multilegged robots is presented in \cite{silva2008}, with a set of performance indices described for optimisation. 

Previous works in optimising the locomotion parameter space have focused on tuning parameters for a single physical embodiment of the robot. Both learning \cite{Heijmink2017} and non-learning \cite{Bjelonic2017} based methods have been used to optimise for energy efficient locomotion. The focus of this work is to expand the locomotion parameter search into the unexplored area of optimising locomotion efficiency for different physical configurations of an over-actuated hexapod robot, along with the traditional locomotion parameters (step frequency, stride length and gait type). While the study is not an exhaustive search of this space, the hardware-based experimental results provide insight into this unique space.

Bullet was designed to overcome locomotion workspace limitations of Weaver \cite{Bjelonic2016} and is capable of switching between insectoid and mammalian leg configurations. In addition to statically stable gaits, we also evaluate the performance of Bullet when using the dynamic gait bipod-B introduced in \cite{Ramdya2017}. The bipod-B gait for hexapods provides a fast and energy optimal gait when foot tips have low traction. With experimental results, we show that the mammalian configuration outperforms the insectoid configuration in energy efficiency at the expense of stability at low step frequencies and maximum body velocity.

\begin{table}[b!]
\vspace{-0.5cm}
    \caption{Hardware Specifications of Bullet. A and B denotes the two kinematic arrangements presented.}  
    \vspace{-0.5cm}

    \label{table:specs}
        \begin{center}
    \begin{tabularx}{3.3 in}{@{} c Y @{}}
    \toprule
    Type & Description \\
    \midrule
    General & Mass (without battery): 8.41\,kg (A), 9.47\,kg (B) \\ 
            & Mass (with battery): 10.07\,kg (A), 11.13\,kg (B) \\
            & Dimensions (Body): $L_{B}$ = 500\,mm x $W_{B}$ = 280\,mm \\
    Servomotors & 30 $\times$ Dynamixel MX-106 \\
    Power supply & Motors: 7-cell LiPo battery (25.9\,V, 5000\,mAh) \\
                 & Computer: 4-cell LiPo battery (14.8\,V, 3300\,mAh) \\
    Computer & Intel i7 NUC PC (16\,GB RAM) running ROS on Ubuntu 16.04 \\ 
    Sensors & IMU (Microstrain GX5 - 100\,Hz) \\
    & Arduino based power monitor (90\,Hz)\\
    \bottomrule
    \end{tabularx}
    \end{center}
\end{table}

\section{Robot Platform}
\label{sec:robotplatform}
Bullet is a versatile 30 DOF hexapod robot designed for traversing rough terrain. Bullet's body length has been extended compared to Weaver's, with legs offset further and perpendicular to the body to allow for the legs to be folded beside the body for a mammalian configuration without limiting the workspace with self-collisions. Fig.~\ref{fig:stancesAB} illustrates Bullet in insectoid and mammalian configurations.
Bullet is capable of un-tethered remote operation with on-board batteries, computer and sensors (described in Table \ref{table:specs}).

\subsection{Leg Kinematic Arrangement and Workspace}
\label{sec:legkinematicworkspace}

\begin{figure}[!t]
    \centering
    \includegraphics[width=85mm]{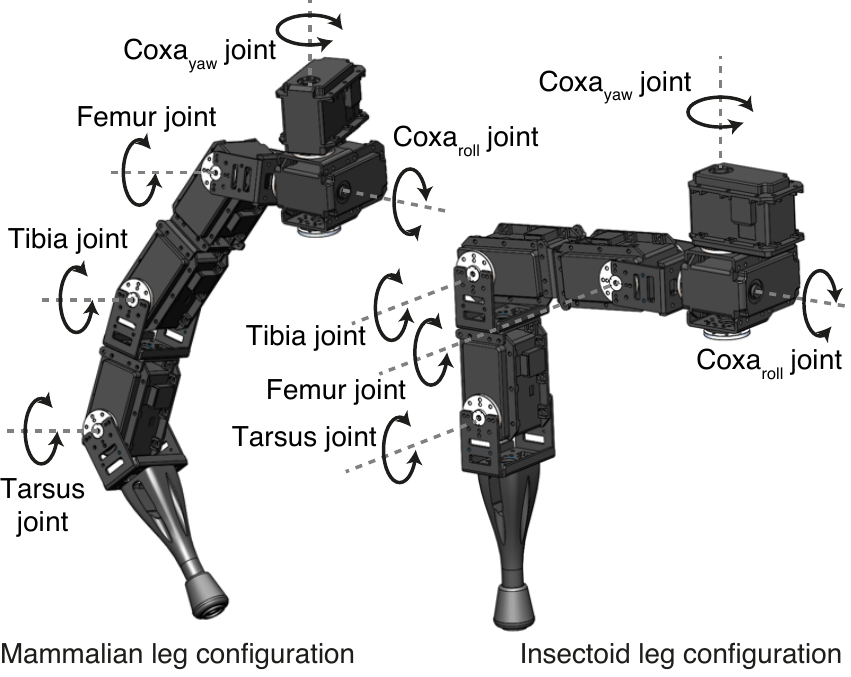}
    \caption{Bullet's leg joint configuration in both mammalian (left) and insectoid (right) configurations in kinematic arrangement A. 
    }
    \label{fig:insect_mammalian_leg_iso}
\end{figure}

\begin{figure}[!t]
    \centering
    \includegraphics[width=85mm]{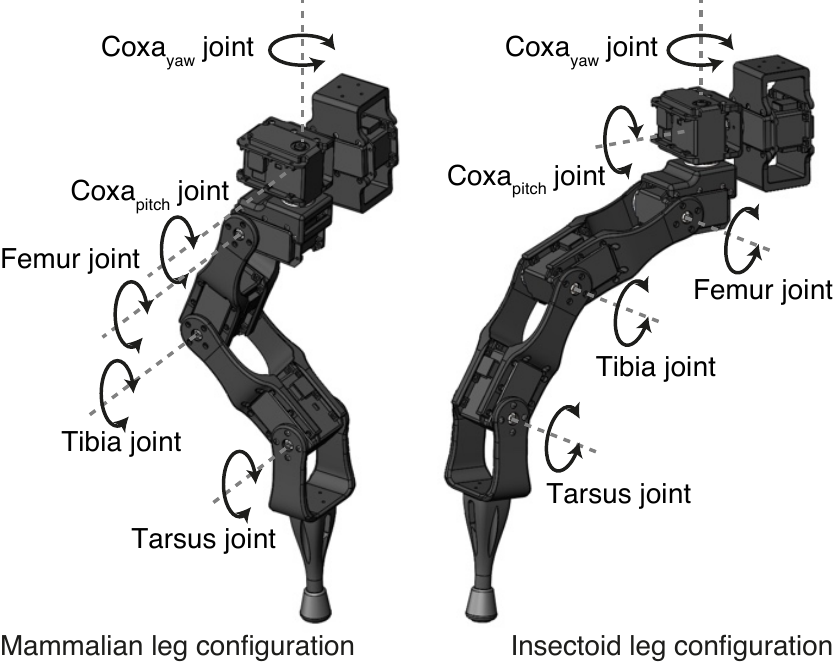}
    \caption{Bullet's leg joint configuration in both mammalian (left) and insectoid (right) configurations in kinematic arrangement B. 
    }
    \label{fig:insect_mammalian_leg_iso_new}
\end{figure}

Two different leg kinematic arrangements were designed through analysis of workspace and joint limits. The original kinematic arrangement (arrangement A), first introduced in\cite{Bjelonic2017}, consists of a yaw-roll-pitch-pitch-pitch kinematic arrangement. This arrangement is compared with a new pitch-yaw-pitch-pitch-pitch kinematic arrangement (arrangement B) with extended link lengths. These joints are named coxa$_\textrm{yaw}$, coxa$_\textrm{roll}$, femur, tibia and tarsus respectively for the former and coxa$_\textrm{pitch}$, coxa$_\textrm{yaw}$, femur, tibia and tarsus respectively for the latter. Arrangement B increases total leg length from 340\,mm to 466\,mm and improves joint ranges by up to 20\degree  through reduction of self collision.  The extended link lengths were set as the minimum length required for the greater joint limits, to lessen the impact of flex and increased motor torques. The change improves the 3D workspace, increases the joint limits and allows the robot to operate inverted (mirrored about the $xy$ plane).
Fig.~\ref{fig:insect_mammalian_leg_iso} and \ref{fig:insect_mammalian_leg_iso_new} illustrates these leg kinematic arrangements in mammalian and insectoid configuration.

Fig.~\ref{fig:workspaces} illustrates the different workspaces for each leg configuration and kinematic arrangement with the values outlined in Table \ref{table:workspaces}. The workspace area is calculated per leg and the stride length is taken as the forwards (robot's $x$-axis) distance of the workspace assuming forward motion.

\begin{table}[b!]
\vspace{-0.5cm}
    \caption{Calculated Workspace of Different Configurations and Arrangements.}  
    \vspace{-0.5cm}
    \label{table:workspaces}
    \begin{center}
    
    \begin{tabular*}{\columnwidth}{@{\extracolsep{\fill}}lllcc}
    \toprule
    Config. & Arrange. & Area ({m$^{2}$}) & Stride Len. (m)\\
    \midrule
    Mammalian & A & 0.049 & 0.207 \\
     & B & 0.063 & 0.210 \\
    Insectoid & A & 0.046 & 0.287 \\
     & B & 0.060 & 0.348 \\
    \bottomrule
    \end{tabular*}
    \end{center}
\end{table}

\subsection{Leg Configuration}
Bullet's unique five DOF legs and body shape allows for self-actuated switching of the leg configuration between mammalian and insectoid without external intervention, different to \cite{Sanz2012}. In the insectoid configuration, five DOFs are available to position and orientate the leg tarsus. To solve for the foot tip position of the over-actuated leg, an additional constraint is used to solve the inverse kinematics problem. The foot tip orientation is constrained to return to the original stance orientation for consistent foot tip touchdown. During the swing phase, the IK solver is able to utilise all joints for motion, following the algorithms outlined in Section~\ref{sec:inversekinematics}. This is different to \cite{Bjelonic2017}, where the tarsus and coxa$_\textrm{roll}$ joints were constrained by the inclination of the robot.

 \begin{figure}[!t]
    \centering
    \includegraphics[width=85mm]{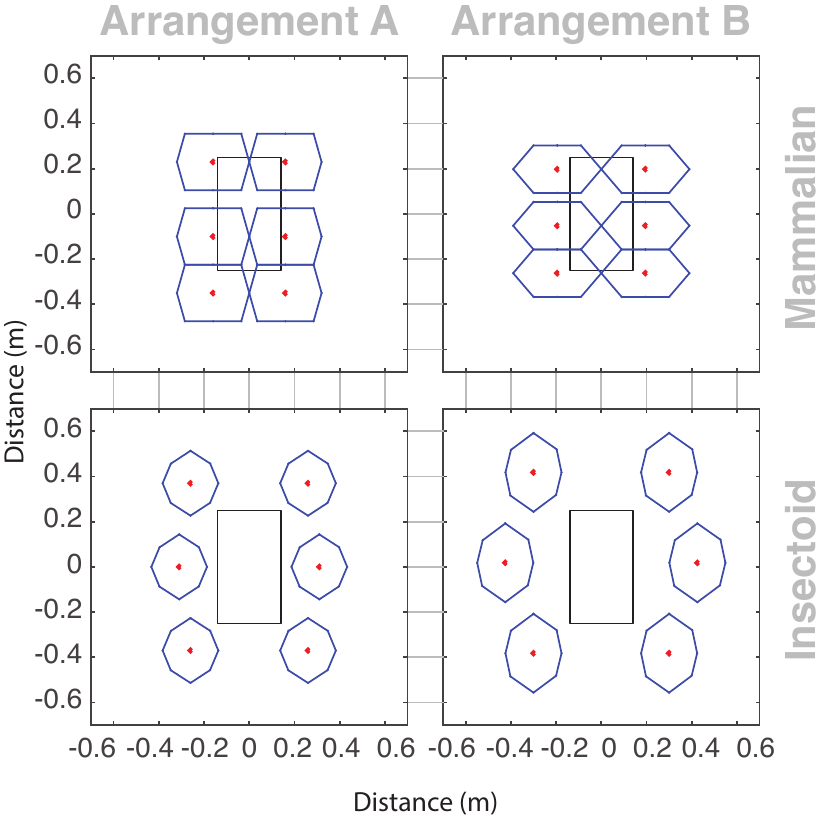}
    \caption{Calculated workspace for the different configurations and arrangements. The rectangle represents the robot body with the robot coordinate frame's origin at its centre. The red dots represent the foot tip locations.
    }
    \label{fig:workspaces}
\end{figure}

To transform Bullet into the mammalian configuration, the coxa$_\textrm{yaw}$ joints are commanded and locked to fixed positions for kinematic arrangement A. The direction of the leg orientation (either forwards or backwards) is controlled through the coxa$_\textrm{yaw}$ joint rotation. For arrangement A, the resulting four DOF leg contains the coxa$_\textrm{roll}$ joint for direction and the three pitch (femur, tibia and tarsus) joints for propulsion. In arrangement B, the coxa$_\textrm{yaw}$ joint is limited, but not locked. This arrangement utilises the coxa$_\textrm{yaw}$ joint for direction (by bending the leg inwards or outwards) and coxa$_\textrm{pitch}$, femur, tibia and tarsus joints for propulsion. Similar to the insectoid configuration, the leg is over-actuated, with the foot tip orientation constrained during touch down. 

For mammalian configuration in quadrupeds, research has found marginal differences in stability depending on the elbows and knees pointing forwards or backwards \cite{Meek2008}. In hexapods, the leg orientation of choice is front leg forwards, and middle and rear legs backwards \cite{Sanz2012}, \cite{Jin2011}. This orientation was implemented on kinematic arrangement A without success on Bullet. This was due to stability issues when walking in tripod gait caused by Bullet's centre of mass and the limited calculated leg workspace. The workspace was limited due to the femur and tibia joints reaching their limits. Through analysis of the workspace and stability while walking, the legs backwards configuration was found to be the most stable for walking while maximising the workspace that borders with the adjacent legs. 
Due to the legs bending backwards, the rear two sets of legs were required to be further behind the body to keep the centre of mass within the support polygon, causing the asymmetrical leg positions for the mammalian configuration as shown in Fig.~\ref{fig:workspaces}. Leg arrangement B follows the same orientation for consistency.

\subsection{Parameter Space}
The parameter search space is defined by physical and locomotion parameters. The physical parameters encapsulate the leg configuration (mammalian and insectoid) and kinematic arrangement (A: yaw-roll-pitch-pitch-pitch or B: pitch-yaw-pitch-pitch-pitch). From these physical parameters, the locomotion parameters of: gait type, step frequency and stride length are optimised. These parameters are summarised as:
\begin{equation}
    \Pi =  C \times A \times G \times L_{s} \times F_{s}
    \label{eq:paramspace}
\end{equation}
where the sets $C = \left \{\textrm{Mammalian, Insectoid} \right \}$, $A = \left \{ \textrm{A, B} \right \}$, $G = \left \{ \textrm{tripod, bipod} \right \}$, $L_{s} = \left \{60,75,90,100 \right \}$ as a percentage of maximum stride length and $F_{s} = \left \{0.4,0.6,0.8,1.0,1.2,1.4,1.6,1.8,2.0,2.2 \right \}$ in Hz. The search for the optimal combinations were limited to motions that were safe for the robot, with some combinations not investigated.


\section{Experiments}
\label{sec:experiments}
The performance of Bullet was tested across flat terrain {to evaluate the effectiveness of the different configurations.}

\subsection{{Performance Metrics}}
The power consumption and joint torques were compared in stance (standing stationary) and while walking on flat terrain. The dimensionless energetic cost of transport ($CoT$) is a performance metric used to compare locomotion of animals \cite{Tucker1975} and also wheeled and legged robots \cite{Bjelonic2016, Kitano2016, Seok2013}. The overall cost of transport ($\overline{CoT}$) over a travelled distance is given by:
\begin{equation} \label{eq:COT}
\overline{CoT} = \frac{\frac{1}{n} \sum\limits_{i=1}^{n} U_i I_i}{mg \frac{\Delta x}{\Delta t}}
\end{equation}
where $U$ is the power supply voltage ($V$), $I$ is the instantaneous power supply current draw ($A$), $n$ is the total number of data points, $m$ is the mass ($kg$), $g$ is the gravitational acceleration ($ms^{-2}$) and $\Delta t$ is the time needed in seconds to travel distance $\Delta x$ ($m$). Herein, $CoT$ refers to the overall $\overline{CoT}$.

The $CoT$ depends on the velocity of the robot \cite{Tucker1975}, where consumption is high at low and high speeds, with a local minimum value at a particular speed. The desired body velocity for the controller is governed by (\ref{eq:velocity}), assuming perfect conditions, ignoring slippage, robot model tolerances, leg link flex and motor errors. Thus, a higher desired velocity can be achieved through increasing either the step frequency or stride length parameter. To account for non-perfect conditions, the body velocity was tracked externally for $CoT$ calculations.

To evaluate the effectiveness of the different configurations, the tripod gait, known for its stability and speed, was compared between mammalian and insectoid at different speeds. Other statically stable gaits such as amble, wave or ripple were not tested as previous works have shown tripod is optimal on flat terrain. The insectoid bipod-B gait \cite{Ramdya2017}, was also compared. OpenSHC allows for parameters, listed in Table~\ref{table:SHCparams}, to be set. These values were heuristically tuned to optimise for energy efficiency using the rules listed in \cite{Sanz2012}.

\begin{table}[b!]
\vspace{-0.5cm}
    \caption{Controller Parameters for Insectoid and Mammalian}  
    \vspace{-0.5cm}
    \label{table:SHCparams}
    \begin{center}
        \begin{tabular*}{\columnwidth}{@{\extracolsep{\fill}}lccc}
    
    \toprule
    Config. & Arrange. & Stride height (m) &  Body height (m)  \\
    \midrule
    Mammalian & A & 0.12 & $H_{m}$ = 0.30 \\
              & B & 0.08 & $H'_{m}$ = 0.30 \\
    Insectoid & A & 0.15 & $H_{i}$ = 0.20 \\
              & B & 0.10 & $H'_{i}$ = 0.25 \\
    \bottomrule
    \end{tabular*}
    \end{center}
\end{table}

\subsection{Experimental Setup}
The power consumption and energy efficiency for the different leg configurations were tested in four scenarios. These scenarios break down the different terms that contribute to power consumption and are:
\begin{itemize}
    \item SA - Power consumed in stance elevated in the air (robot lifted up, no ground contact),
    \item WA - Power consumed walking elevated in the air (robot lifted up, no ground contact),
    \item SG - Power consumed in stance on the ground, and
    \item WG - Power consumed walking on the ground.
\end{itemize}
These scenarios follow the experiments set out in \cite{Sanz2012} where: scenario SA (stance air) provides the power consumed by the motor power circuitry; scenario WA (walk air) provides the power consumed for leg movements; scenario SG (stance ground) for supporting the robot's body weight; and scenario WG (walk ground) for overall power consumption. The robot was elevated on a stand with legs in mid-air for scenarios SA and WA. The robot was placed on the ground and walked once in place to measure the power consumption when stationary in scenario SG. Bullet was analysed traversing flat ground in scenario WG. A flat, slightly inclined 3\,m straight track was marked for the robot. For the calculation of ${CoT}$ in (\ref{eq:COT}), power consumption ($P=UI$) was logged with a power monitor at 90\,Hz and velocity $v$ was measured with a total station (Leica TS12) at approximately 4\,Hz. The $CoT$ is based on the power draw of the motors, including mechanical energy, heat dissipation and friction, while the computer and sensors power consumption is assumed to be constant across the different configurations and arrangements, and are powered on a separate power supply. Bullet was tested using a tether to an external power supply and control computer, with the internal batteries and computer on-board to simulate the real weight of the robot. The power supply was set at 25\,V and the control computer is an i7 laptop with 8\,GB of RAM.

\section{Results}
\label{sec:results}

The metric used to compare scenarios SA, WA and SG is power consumption with results outlined in Table~\ref{table:stancepower}, while $CoT$ is used for scenario WG. The power consumed in stance for both kinematic arrangements of mammalian is lower than their corresponding insectoid. This validates the theory of less torque on the motors with the legs underneath as the force vectors act through the motor shaft.
The difference in power between the air and ground shows the additional power required to support the robot's weight. 
While arrangement A consumes less power than B in stance while in the air, on the ground arrangement B consumes less. This suggests that the joints in arrangement B require less torque to keep the body up. The walking in air comparison is for step frequency 1\,Hz at 100\% stride length. The results show the insectoid bipod consuming the highest amount of energy for leg motion, while in mammalian, arrangement B uses less power than arrangement A. 

\begin{table}[b!]
\vspace{-0.5cm}
    \caption{Average Power Consumption (W)} 
    \vspace{-0.5cm}
    \label{table:stancepower}
    \begin{center}
    \begin{tabular}{l@{}c@{}cccc}
    \toprule
     Config. & Arrange. & Gait & Stance air & Walk air & Stance ground \\
    \midrule
    Mammalian & A & Tripod & 27.7 & 51.0 & 48.6 \\
              & B & Tripod & 30.4 & 47.1 & 46.7 \\
    Insectoid & A & Tripod & 30.4 & 61.9 & 58.3 \\
              & A & Bipod  & 30.4 & 72.3 & 58.3\\
              & B & Tripod & 31.1 & 62.9 & 56.0 \\
    \bottomrule
    \end{tabular}
    \end{center}
\end{table}

\begin{figure}[!t]
    \centering
    \includegraphics[width=85mm]{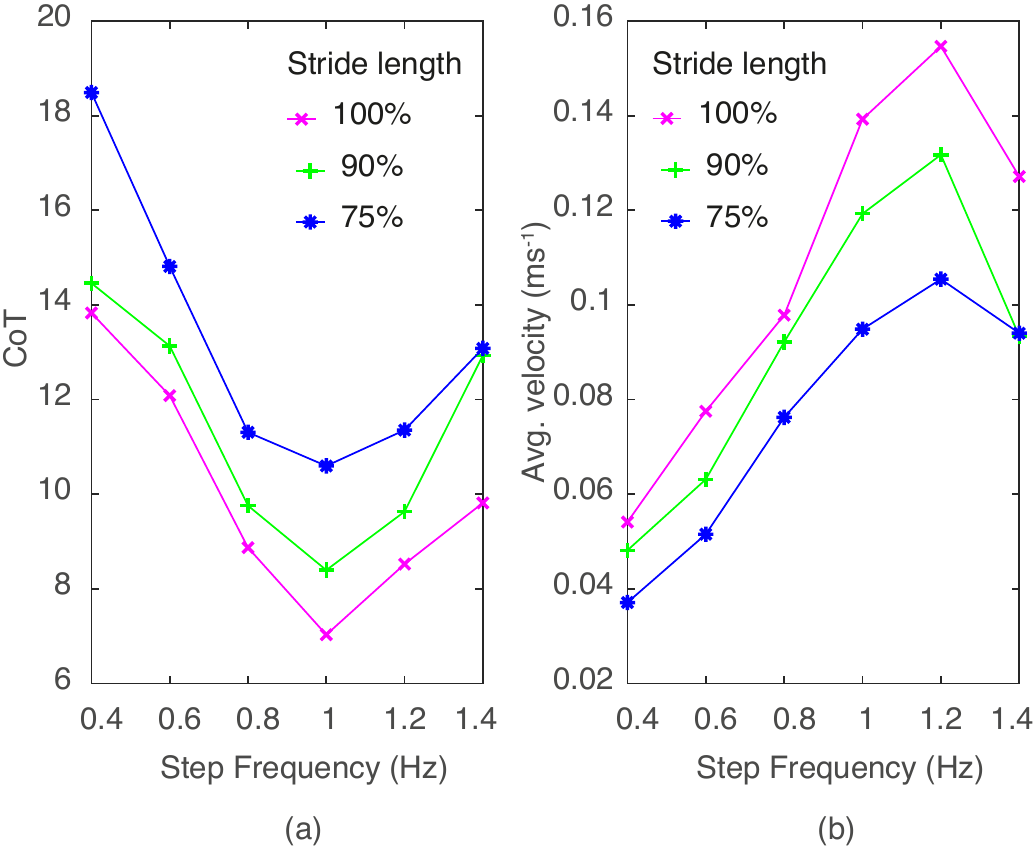}
    \caption{$CoT$ for arrangement A mammalian tripod gait at various step frequencies and stride lengths.}

    \label{fig:mt}
\end{figure}

An analysis of arrangement A mammalian configuration at different step frequency and stride length compared to the $CoT$ is shown in Fig.~\ref{fig:mt}. The stride length, given as a percentage of the available locomotion workspace, shows that a higher length for a given step frequency is more energy efficient. The minimum value at 1\,Hz shows the optimal step frequency and stride length for energy efficient locomotion. An increase of step frequency above 1\,Hz increases the power usage greatly. At a step frequency of 1.4\,Hz, the foot tips slipped considerably, causing the tracked velocity to be lower than that of step frequency 1.2\,Hz; even though the desired body velocity is greater.

 \begin{figure}[!t]
    \centering
    \includegraphics[width=85mm]{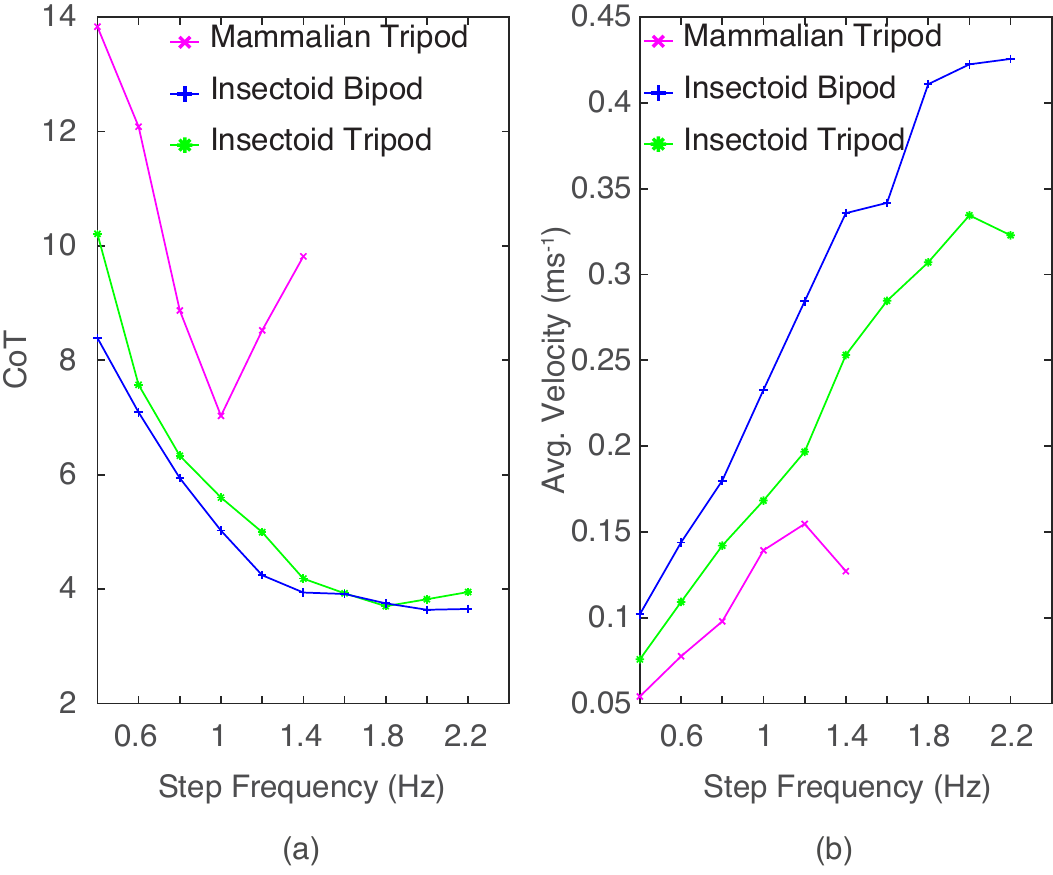}
    \caption{$CoT$ for insectoid bipod and tripod, and mammalian tripod gait at various step frequencies at 100\% stride length for arrangement A.}
    \label{fig:itb}
\end{figure}

 \begin{figure}[!t]
    \centering
    \includegraphics[width=85mm]{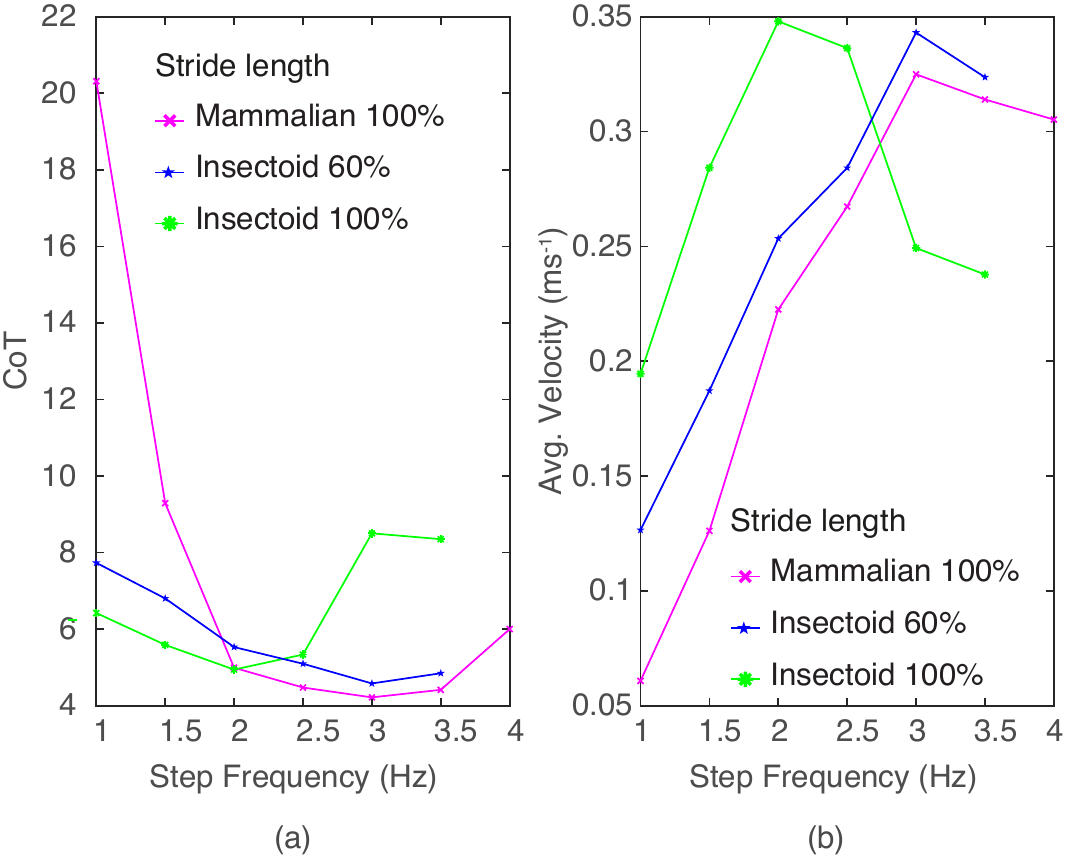}
    \caption{ $CoT$ for mammalian and insectoid tripod gait at various step frequencies and stride lengths for arrangement B. Insectoid stride length 60\% was limited to be the same metric length as mammalian 100\% stride length.}
    \label{fig:newleg}
\end{figure}

The insectoid configuration was analysed with the tripod and bipod gait in arrangement A. A stride length of 100\% was used in tests as results in mammalian showed a stride length of 100\% is the most efficient across all step frequencies tested. Fig.~\ref{fig:itb} compares the $CoT$ and average velocity as the step frequency increases. Similar to mammalian configuration, tripod gait has a minimum $CoT$, occurring at 1.8\,Hz. An increase of step frequency above 1.8\,Hz increases the $CoT$. The bipod gait does not exhibit this minimum, instead having diminished reductions for step frequencies about 2\,Hz. The bipod gait performed equally or better than the tripod gait at all step frequencies. For step frequencies above 2\,Hz, both insectoid gaits cause considerable clamping of the joint angles by OpenSHC. This caused body instability and the foot tip to no longer track the trajectory. Step frequencies above 2.2\,Hz were not tested to prevent hardware damage.

For arrangement B, tests were conducted to compare the mammalian and insectoid configuration given the same workspace across various step frequencies. The mammalian configuration was tested at 100\% stride length. This was compared with insectoid at 60.27\% of its maximum workspace to have the same metric stride length as mammalian. Fig.~\ref{fig:newleg} outlines the $CoT$ and average velocity as the step frequency increases. Insectoid at 100\% stride length was included for comparison. At low step frequencies, the mammalian configuration was observed to have body oscillation and foot tip slip, resulting in the high $CoT$. At step frequency 2\,Hz and above, where the body oscillations were minimal, the mammalian configuration has comparable or better performance than both insectoid stride lengths. Given the same stride length, mammalian configuration consistently has a lower $CoT$ than insectoid across step frequencies above 2\,Hz (Fig.~\ref{fig:newleg}a). However, the actual robot velocity is consistently lower than insectoid (Fig.~\ref{fig:newleg}b). The local minimum value for $CoT$ occurs at step frequency 3\,Hz for both configurations, suggesting it is the step frequency and stride length that affects the local minimum $CoT$ and not the leg configuration. 

The comparison of 100\% stride length between the two configurations show the advantages and disadvantages of each. The insectoid configuration can achieve a higher maximum velocity at lower step frequencies and is also more stable. However, it causes the joints to reach limits at a lower frequency (observed at 2\,Hz compared to mammalian at 3.5\,Hz). The drop in velocity in insectoid at high step frequencies is caused by joint velocity clamping while executing the desired controller trajectory. As the weight of the robot is constant across the different trials, the $CoT$ is effectively the relative power required to traverse a set distance, and can be directly used to compare energy efficiency of the different configurations.

\begin{figure}[!t]
    \centering
    \includegraphics[width=85mm]{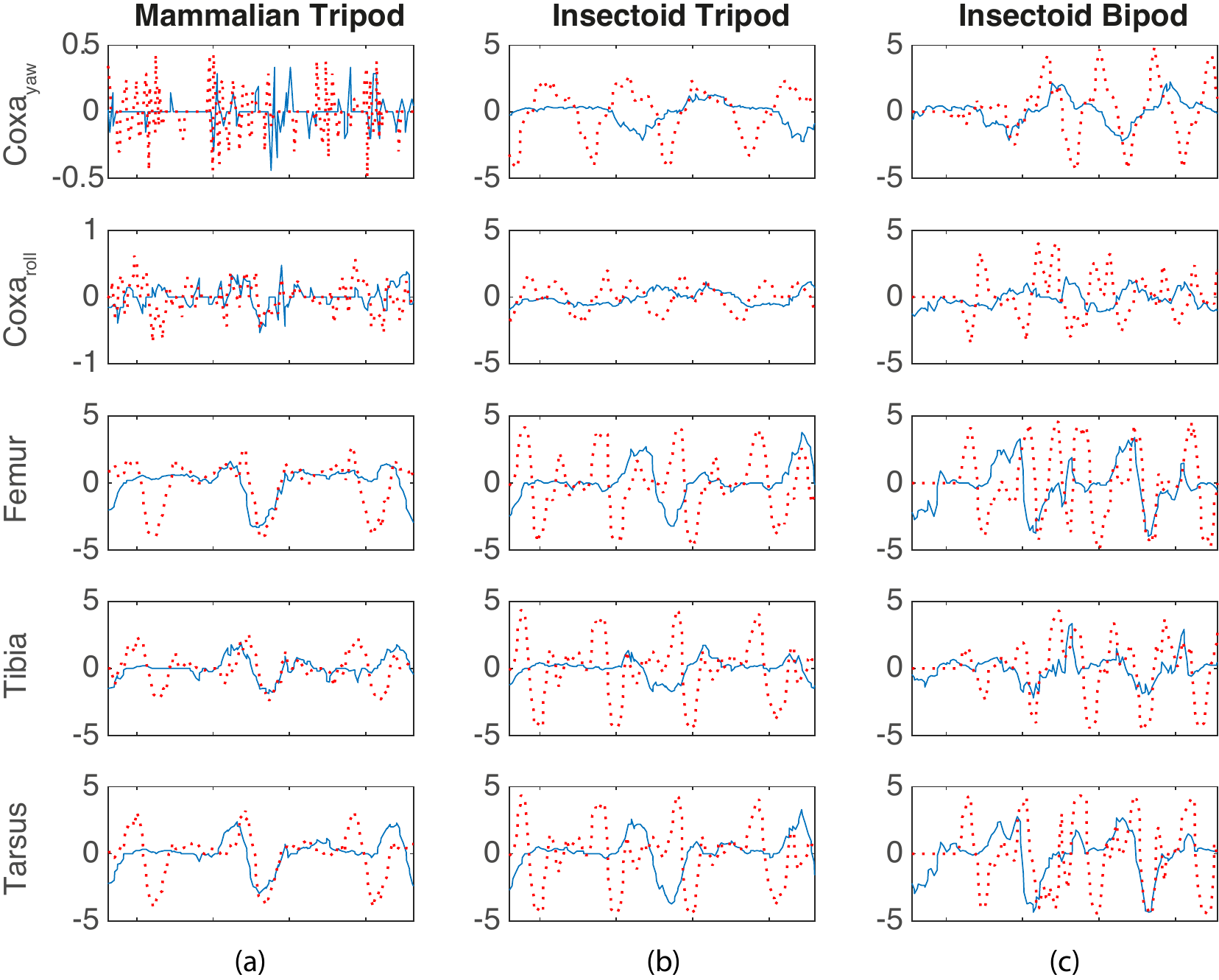}
    \caption{Velocity ($rads^{-1}$) profile of the joints in the front right leg for arrangement A. The solid blue line denotes a step frequency of 1\,Hz at 100\% stride length. The dotted red line denotes the maximum step frequency at 100\% stride length for each configuration (1.4\,Hz, 2.2\,Hz and 2.2\,Hz respectively). The $x$-axis denotes two periods of the step frequency at 1\,Hz.
    }

    \label{fig:joint_vel}
\end{figure}

\begin{figure}[!t]
    \centering
    \includegraphics[width=85mm]{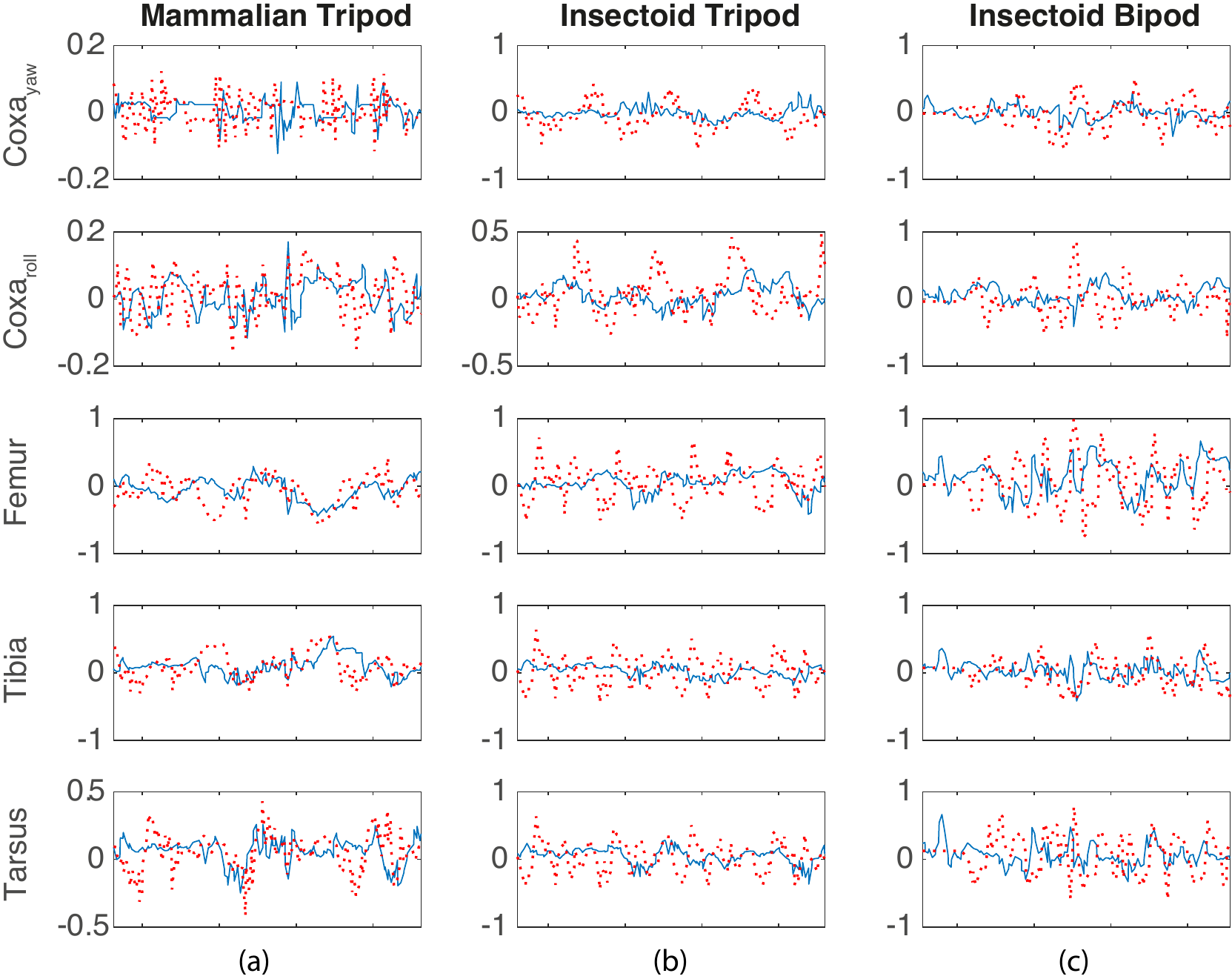}
    \caption{Load profile (dimensionless) of the joints in the front right leg for arrangement A. The solid blue line denotes a step frequency of 1\,Hz at 100\% stride length. The dotted red line denotes the maximum step frequency at 100\% stride length for each configuration (1.4\,Hz, 2.2\,Hz and 2.2\,Hz respectively). The $x$-axis denotes two periods of the step frequency at 1\,Hz.
    }
    \label{fig:joint_load}
\end{figure}

\section{Discussion}
\label{sec:discussion}
 The use of an over-actuated leg to achieve switching between mammalian and insectoid configuration was explored for the first time. To achieve the mammalian configuration, the coxa$_\textrm{yaw}$ joint motors were commanded to set positions, with torque holding those positions. This was reflected by the small velocity and load on the joint illustrated in Fig.~\ref{fig:joint_vel}a and \ref{fig:joint_load}a respectively. Comparing the load on the joints across the different gaits (Fig.~\ref{fig:joint_load} blue line), the mammalian tripod gait had lower torques on all the joints. The results for the stance power consumption experiments (scenarios SA and SG) confirmed the theory in \cite{Todd1985} and also the experimental data in \cite{Sanz2012} with the mammalian configuration using less energy than insectoid when standing.

Arrangement A mammalian configuration consumed less power than insectoid in stance (SG) and leg movement (WG), but was not the most efficient in locomotion due to the smaller locomotion workspace and joint angle limits. The robot in insectoid configuration was capable of covering larger distances in the same number of steps, negating the higher power required to support its weight and for locomotion. With the joint limit improvements in arrangement B, the mammalian configuration achieved lower $CoT$ than insectoid. The mammalian configuration was optimal if the least amount of power to traverse an area is required. Arrangement B mammalian did not reach joint velocity limits until higher step frequencies as the pitch-yaw-pitch-pitch-pitch kinematic arrangement affords four pitch joints to propel the robot forwards.

Analysis of the insectoid tripod and bipod gait showed a lower $CoT$ for the latter. Although the bipod gait consumed more power, it was able to walk faster than tripod at the same step frequency. This supports the hardware experiments conducted in \cite{Ramdya2017}. On average, the bipod gait was 30\% faster than the equivalent tripod gait, similar to the 25\% increase reported in \cite{Ramdya2017}. Between the two arrangements and across different gaits, arrangement A bipod had the lowest $CoT$. The results showed improved locomotion efficiency is achievable with OpenSHC, providing direction for future works in the absence of sophisticated dynamic controllers.

The versatility of OpenSHC was highlighted on experiments conducted on a versatile hexapod capable of switching between different configurations. We show that this switching ability has advantages over traditional fixed configuration robots in autonomous navigation of terrain. When traversing mild terrains, where stability and speed is of less importance, the mammalian configuration consumes less power. For terrain or scenarios where stability or speed is required (such as rough terrain or speed critical tasks), the robot can morph into insectoid configuration. This ability to select the optimal configuration provides the robot an advantage on locomotion efficiency for real world scenarios. Having an over-actuated leg design allows for the additional ability to selectively use particular motors for locomotion and to switch between the different configurations. While a single DOF is unused in the mammalian configuration, setting the motor to a fixed position results in the robot using less power overall. The results highlight the advantages of a versatile hexapod running OpenSHC, capable of selecting between configurations and step frequencies, based on energy efficiency (mammalian) or stability and speed (insectoid) requirements.
\section{Conclusions}
\label{sec:conclusions}

{This paper presented OpenSHC, a versatile controller capable of generating smooth trajectories for quasi-static legged robots. With many customisable parameters, the controller can be configured for new robot platforms with various physical characteristics. OpenSHC provides the building blocks to extend the capabilities of multilegged robot platforms through a modular hierarchical architecture. By providing `out of the box' functionality for locomotion in mild to rough terrain, OpenSHC allows researchers to focus on higher level areas such as autonomy or application specific motions that would increase the versatility of legged robots. With additional sensors, robots would be capable of traversing in more complex and confined environments, something that quasi-static multilegged platforms would be ideal for. Another area of research that OpenSHC supports is in leg manipulation. For robot platforms with greater than 3\,DOF legs, OpenSHC allows for manual position and orientation control of the foot tip, allowing attachments such as grippers to manipulate the environment. Additionally, with its seamless integration with rviz and Gazebo simulation environments, OpenSHC provides a way for researchers to design, test and optimise new legged robots in simulation and make informed choices about hardware based on application requirements.}

\appendices

\section{{Forward Kinematics Derivation}}
\label{sec:FKderivation}

The DH parameters $\theta_{i}$, $d_{i}$, $a_{i}$ and $\alpha_{i}$ used in (\ref{eq:DHTransform}) from Section~\ref{sec:forwardkinematics} are represented as transformation matrices using the shorthand $c_{x}$ and $s_{x}$ to denote $\cos(x)$ and $\sin(x)$ respectively.

The transformation matrix for the DH parameter $\theta_{i}$ is given by:
\begin{equation}
    Rot_{z,\theta_{i}} = 
\begin{bmatrix} 
            c_{\theta_{i}} & -s_{\theta_{i}} & 0 & 0 \\
            s_{\theta_{i}} & c_{\theta_{i}} & 0 & 0 \\
            0 & 0 & 1 & 0 \\
            0 & 0 & 0 & 1
            \end{bmatrix}
\end{equation}

The transformation matrix for the DH parameter $d_{i}$ is given by:
\begin{equation}
     Trans_{z,d_{i}}= \begin{bmatrix} 
            1 & 0 & 0 & 0 \\
            0 & 1 & 0 & 0 \\
            0 & 0 & 1 & d_{i} \\
            0 & 0 & 0 & 1
            \end{bmatrix}
\end{equation}

The transformation matrix for the DH parameter $a_{i}$ is given by:
\begin{equation}
    Trans_{x,a_{i}}= \begin{bmatrix} 
            1 & 0 & 0 & a_{i} \\
            0 & 1 & 0 & 0 \\
            0 & 0 & 1 & 0 \\
            0 & 0 & 0 & 1
            \end{bmatrix}
\end{equation}

The transformation matrix for the DH parameter $\alpha_{i}$ is given by:
\begin{equation}
    Rot_{x,\alpha_{i}} = \begin{bmatrix} 
            1 & 0 & 0 & 0 \\
            0 & c_{\alpha_{i}} & -s_{\alpha_{i}} & 0 \\
            0 & s_{\alpha_{i}} & c_{\alpha_{i}} & 0 \\
            0 & 0 & 0 & 1
            \end{bmatrix}
\end{equation}

\section{{Inverse Kinematics Derivation}}
\label{sec:IKderivation}

The IK for the change in joint angles for a given desired end effector position is calculated using the Jacobian. The Jacobian matrix $J$ is defined as:
\begin{equation}
    J \left( \theta \right) =\left(\frac{\partial \text{s}_i}{\partial \theta_j}\right)_{i,j}
\end{equation}
where $\text{s}_i$ is the \textit{end effector} of the link which is affected by joint $\theta_j$.
The linear velocity component of the Jacobian can be calculated by:
\begin{equation}
    \frac{\partial s_i}{\partial \theta_j} = \text{v}_j \times \left(\text{s}_i - \text{p}_j\right)
\end{equation}
where $\text{v}_j$ is the unit vector along the current axis of rotation and $\text{p}_j$ is the position of the $j^{\text{th}}$ joint.
For incremental iterations of the end effector, a change in the end effector can be approximated by:
\begin{equation}
    \Delta \vec{\mathbf{s}} \approx J \Delta \theta.
\end{equation}

Thus, using the Levenberg-Marquardt method, also known as the damped least squares method, $\Delta \theta$ can be solved by:
\begin{equation}
    \Delta \theta \approx J^T\left( J J^T + \lambda^2I\right)^{-1}  \Delta \vec{\mathbf{s}}
\end{equation}
where $\lambda > 0\in \mathbb{R}$.
Using this approximation and compact notation, we get:
\begin{equation}
    \Delta \theta = J^T\left( J J^T + \lambda^2I\right)^{-1}  \Delta \vec{\mathbf{s}} = \mathbf{Z} \Delta \vec{\mathbf{s}}
\end{equation}
where $ \mathbf{Z} = J^T\left( J J^T + \lambda^2I\right)^{-1} $.

The cost function to minimise for JLA in OpenSHC is represented by:
\begin{equation}
    \Phi \left (q \right) = \sum_{i=1}^{n}\left [  \frac{q_i-q_{c_i}}{\Delta q_i}\right ]^{2}
    \label{eq:jla1}
\end{equation}
where $q_{c_i}$ is the centre of the joint range $\Delta q_i$ for joint $i$.


\section*{Acknowledgement}
The authors would like to thank Oshada Jayasinghe, Marisa Bucolo, James Brett, Isuru Kalhara, Eranda Tennakoon, Thomas Molnar, Benjamin Wilson, Thomas Lowe, and David Rytz for their support during this project.

\balance
\bibliographystyle{IEEEtran}

\end{document}